\documentclass[runningheads]{llncs}
\usepackage{graphicx}
\usepackage{amsmath,amssymb}
\usepackage{xpatch} 
\usepackage[utf8]{inputenc} 
\usepackage[T1]{fontenc}
\usepackage{cite}
\usepackage{hyperref}       
\usepackage{url}            
\usepackage{booktabs}       
\usepackage{amsfonts}       
\usepackage{nicefrac}       
\usepackage{microtype}      %
\usepackage{xcolor}         
\usepackage{adjustbox}
\usepackage{amsmath}
\usepackage{subcaption}
\usepackage{sidecap}
\usepackage{multirow}
\sidecaptionvpos{figure}{t}
\usepackage{color}
\usepackage{colortbl}
\definecolor{Gray}{gray}{0.90}
\newcommand{\etal}{\textit{et al}}
\newcommand{\eg}{\textit{e.g}}

\begin{document}

\title{Self-Distilled Vision Transformer for Domain Generalization} 
\titlerunning{ERM-SDViT} 


\author{Maryam Sultana\inst{1, 2} \and
Muzammal Naseer\inst{1, 3} \and
Muhammad Haris Khan\inst{1} \and
Salman Khan\inst{1, 3}\and
Fahad Shahbaz Khan\inst{1, 4}}

\authorrunning{M. Sultana et al.}

\institute{Mohamed Bin Zayed University of AI,
UAE,
\and
VAIL, Oxford Brookes University, UK,
\and
Australian National University,
AU,
\and
Linköping University,
Sweden,\\
\email{maryam.sultana, muzammal.naseer, muhammad.haris, salman.khan, fahad.khan, @mbzuai.ac.ae}}

\maketitle

\begin{abstract}
In the recent past, several domain generalization (DG) methods have been proposed, showing encouraging performance, however, almost all of them build on convolutional neural networks (CNNs). There is little to no progress on studying the DG performance of vision transformers (ViTs), which are challenging the supremacy of CNNs on standard benchmarks, often built on i.i.d assumption. This renders the real-world deployment of ViTs doubtful.
In this paper, we attempt to explore ViTs towards addressing the DG problem. Similar to CNNs, ViTs also struggle in out-of-distribution scenarios and the main culprit is overfitting to source domains. Inspired by the modular architecture of ViTs, we propose a simple DG approach for ViTs, coined as \emph{self-distillation for ViTs}. It reduces the overfitting of source domains by easing the learning of input-output mapping problem through curating non-zero entropy supervisory signals for intermediate transformer blocks. Further, it does not introduce any new parameters and can be seamlessly plugged into the modular composition of different ViTs. We empirically demonstrate notable performance gains with different DG baselines and various ViT backbones in five challenging datasets. Moreover, we report favorable performance against recent state-of-the-art DG methods. Our code along with pre-trained models are publicly available at: \textcolor{blue}{\url{https://github.com/maryam089/SDViT}}.

\end{abstract}
\section{Introduction}
Since their inception, transformers have displayed remarkable performance in various natural language processing (NLP) tasks \cite{vaswani2017attention,devlin2018bert,brown2020language}. Owing to their success in NLP, recently, transformer design has been adopted for vision tasks \cite{dosovitskiy2020image}. Since then, we have been witnessing several vision transformer (ViT) models for image recognition \cite{dosovitskiy2020image}, object detection \cite{carion2020end,zhu2020deformable} and semantic segmentation \cite{zheng2021rethinking,wang2021pyramid}. ViTs are intrinsically different in design compared to convolution neural networks (CNNs), since they lack explicit inductive biases such as spatial connectivity and translation equivariance. They process (input) image as a sequence of patches that is enhanced via successive transformer blocks (comprised of self-attention mechanisms), thereby allowing the network to model the relationship between any parts of the image. A useful consequence of such processing is a wide receptive field that facilitates capturing the global context in contrast to a limited receptive field modeled in CNNs.

Many deep learning models are usually deployed in real-world scenarios where the test data is unknown in advance. When their predictions are used for decision making in safety-critical applications, such as medical diagnoses or self-driving cars, an erroneous prediction can lead to dangerous consequences. This typically occurs because there is a distributional gap between the training and testing data. Hence, it is critical for deep learning models to provide reliable predictions that generalize across different domains. Domain generalization (DG) is a problem setting in which data from multiple source domains is leveraged for training to generalize to a new (unseen) domain \cite{muandet2013domain,ghifary2015domain,li2017deeper,carlucci2019domain,Gulrajani2021InSO,huang2020self,khan2021mode,Nam_2021_CVPR,Kim_2021_ICCV,bui2021exploiting}. 
Existing DG methods aim to explicitly reduce domain gap in the feature space \cite{muandet2013domain, ganin2016domain, li2018domain}, learn well-transferable model parameters through meta-learning \cite{li2018learning,dou2019domain, balaji2018metareg,li2019episodic}, propose different data augmentation techniques \cite{shankar2018generalizing,volpi2018generalizing,zhou2020learning,khan2021mode}, or leverage auxiliary tasks \cite{carlucci2019domain,wang2020learning}. Lately, Gulrajani and Lopez-Paz \cite{Gulrajani2021InSO} show that a simple Empirical Risk Minimization (ERM) method obtains favorable performance against previous methods under a fair evaluation protocol termed as  ``Domainbed''.
To our knowledge, almost all aforementioned DG approaches are based on CNNs, and there is little to no work on studying the DG performance of ViTs. So, in effect, despite ViTs demonstrating state-of-the-art performance on some standard benchmarks, often rooted in i.i.d assumption, their real-world deployment remains doubtful. To this end, we attempt to explore ViTs towards addressing the DG problem.

\begin{SCfigure}[][!t]
\caption{\small In-domain (validation) and out-of-domain (target) classification accuracy of ERM-CNN and ERM-ViT in four DG datasets. Similar to ERM-CNN, ERM-ViT also shows performance degradation in out-of-domain scenarios. }
\resizebox{0.5\textwidth}{!}{
  \includegraphics[width=\textwidth]{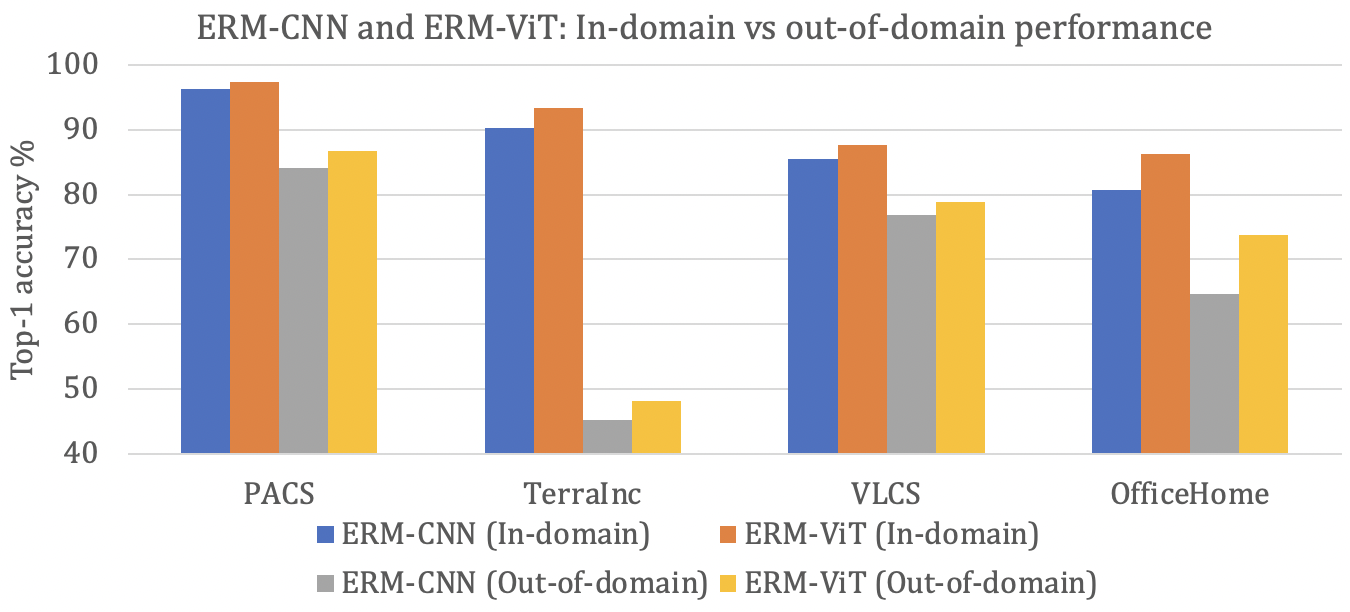}}  \label{fig:in_domain_out_domain_comparison}
\end{SCfigure}

\begin{SCfigure}[][t]
\caption{\small Attention maps from ERM-ViT (top) and ERM-SDViT (bottom) corresponding to images of the target domains in \texttt{PACS}. ERM-ViT is prone to exploit non-generalizable features e.g., background. Whereas ERM-SDViT is capable of learning cross-domain generalizable features e.g., object shape and its semantics.}
\resizebox{0.5\textwidth}{!}{
  \includegraphics[width=\textwidth]{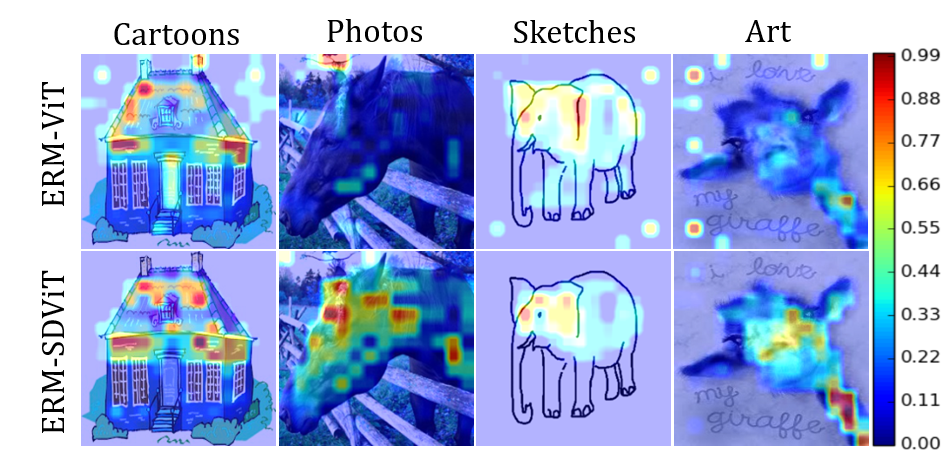}}
   \label{fig:motivation_fig}
\end{SCfigure}

We note that ViT-based ERM (ERM-ViT), similar to its CNN-based counterpart (ERM-CNN), also suffers from performance degradation when facing out-of-distribution (OOD) target domain data (see Fig.~\ref{fig:in_domain_out_domain_comparison}). In the absence of any explicit overfitting prevention mechanism coupled with the one-hot encoded ground-truth supervision, which is essentially zero-entropy signals, as such it is challenging for a simple ERM-ViT model to obtain favorable OOD generalization. Under the hood, since the mapping problem is difficult, the model is prone to inadvertently exploiting non-generalizable, brittle features for making predictions. Fig.~\ref{fig:motivation_fig} visualizes attention maps from ERM-ViT on arbitrary images of four target domains in \texttt{PACS} dataset. ERM-ViT has the tendency to rely on non-object related features such as the background features, which are potentially non-transferable between the source and the target domains.  

Inspired by the modular architecture of ViTs, we propose a light-weight plug-and-play DG approach for ViTs, namely \textit{self-distillation for ViT (SDViT)}. It explicitly encourages the model towards learning generalizable, comprehensive features. 
ViTs process a sequence of input image patches repeatedly by multiple multi-headed self-attention layers, a.k.a transformer blocks \cite{vaswani2017attention}. 
%
These image patches are also known as \textit{patch tokens}. A randomly initialized class token is usually appended to the set of image patches (tokens). This group is then passed through a sequence of transformer blocks followed by the passing of class token through a linear classifier to get final predictions. The class token can learn information that is useful while making a final prediction. So, it can be extracted from the output of each transformer block and can be leveraged to get class-specific logits using the final classifier of the pretrained model \cite{naseer2021improving}. 
Armed with this insight, we propose to transfer the so-called dark knowledge from the final classifier output to the intermediate blocks by developing a self-distillation strategy for ViT (sec.~\ref{subsection:Self-distilled Vision Transformers for Domain Generalization}). It alleviates the overfitting of source domains by moderating the learning of input-output mapping problem via non-zero entropy supervision of intermediate blocks.
We show that improving the intermediate blocks, which are essentially multiple feature pathways, through soft supervision from the final classifier facilitates the model toward learning cross-domain generalizable features (see Fig.~\ref{fig:motivation_fig}). 
Our approach naturally fits into the modular and compositional architecture of different ViTs, and does not introduce any new parameters. As such it adds a minimal training overhead over the baseline.
Extensive experiments have been conducted on five diverse datasets from DomainBed suite \cite{Gulrajani2021InSO}, including \texttt{PACS}, \texttt{VLCS}, \texttt{OfficeHome}, \texttt{TerraIncognita}, and \texttt{DomainNet}. We empirically show better performance across different DG baselines as well as different ViT backbones in all five datasets. Further, we demonstrate competitive performance against the recent state-of-the-art DG methods. With CvT-21 backbone, we obtain an (overall) average accuracy (five datasets) of 68.6\%, thereby outperforming the existing best \cite{cha2021swad} by 1.8\%.

\section{Related Work}
\textbf{Domain Generalization:}
A prevalent motivation of several existing DG methods is to learn the underlying domain-invariant representations from the available source data. The seminal work of Vapnik \etal. \cite{vapnik1999nature} introduced Empirical Risk Minimization (ERM), which minimizes the sum of squared errors across domains. Following this route, we observe several variants aimed at uncovering the domain-invariant features via matching distributions across domains. For instance, Muandet \etal. \cite{muandet2013domain} employed maximum mean discrepancy (MMD) constraint, Ghifary \etal. \cite{ghifary2015domain} proposed a multi-task autoencoder, and Yang \etal. \cite{yang2013multi} used canonical correlation analysis (CCA). Arjovsky \etal. \cite{arjovsky2019invariant} proposed the learning of invariant predictors across various source domains. A few methods used low-rank regularization to extract invariant features e.g., \cite{li2017deeper,xu2014exploiting}.
Meta-learning based methods have also been used as regularizers. Li \etal. \cite{li2019episodic} switched domain-specific feature extractors and classifiers across domains via episodic training. Balaji \etal. \cite{balaji2018metareg} learned a regularization function in an episodic training paradigm. Furthermore, some DG methods masked features via ranking gradients \cite{huang2020self}, utilized auxiliary tasks \cite{carlucci2019domain,wang2020learning}, employed domain-specific masks \cite{chattopadhyay2020learning}, and exploited domain-specific normalizations \cite{seo2019learning}. A few DG approaches proposed contrastive semantic alignment and self-supervised contrastive formulations \cite{motiian2017unified,dou2019domain,Kim_2021_ICCV}. Another class of DG methods employs various data augmentation techniques to improve the diversity of source domains. Shankar \etal.\cite{shankar2018generalizing} proposed Crossgrad training, Volpi \etal. \cite{volpi2018generalizing} imposed wasserstein constraint in semantic space, Zhou \etal. \cite{zhou2020learning} learned a generator to generate new examples, and Khan \etal. \cite{khan2021mode} estimated class-conditional covariance matrices for generating novel source features.
Recently, Gulrajani \etal. \cite{Gulrajani2021InSO} demonstrated that, under a fair evaluation protocol, a simple empirical risk minimization (ERM) method can achieve state-of-the-art DG performance. Cha \etal. \cite{cha2021swad} proposed stochastic weight averaging in a dense manner to achieve flatter minima for DG. We note that, all aforementioned DG methods are based on CNN architecture, however, little to no attention has been paid to investigating the DG performance of ViTs. To this end, we choose to study the performance of ViTs under domain generalization with ERM as a simple, but strong DG baseline. 

\noindent\textbf{Vision transformers:} operate in a hierarchical manner by processing input images as a sequence of non-overlapping patches via the self-attention mechanism. 
Recently, we have seen some ViT-based methods for image classification \cite{dosovitskiy2020image,wu2021cvt,touvron2021training}, object detection \cite{carion2020end,Dai_2021_ICCV}, and semantic segmentation \cite{strudel2021segmenter,lu2021simpler}. 
Dosovitskiy \etal. \cite{dosovitskiy2020image} proposed the first fully functional ViT model for image classification. Despite its promising performance, its adoption remained limited because it requires large-scale datasets for model training and huge computation resources. 
Towards improving data efficiency in ViTs, Touvron \etal. \cite{touvron2021training} developed Data-efficient image Transformer (DeiT); it attains competitive results against the CNN by training only on ImageNet and without leveraging external data. 
Similarly, Yuan \etal. \cite{yuan2021tokens} proposed Tokens-To-Token Vision Transformer (T2T-ViT) strategy. It progressively structurizes the patch tokens in a way that the local structure represented by surrounding tokens can be modeled while reducing the tokens length. 
Furthermore, Wu \etal. \cite{wu2021cvt} proposed a hybrid approach, namely Convolutional Vision Transformer (CvT), by combining the strengths of CNNs and ViTs aimed at improving the performance and robustness of ViTs, while maintaining computational and memory efficiency. 
Recently, Zhang \etal. \cite{zhang2021delving} studied the performance of ViTs under distribution shifts and proposed a generalization-enhanced vision transformer from the outlook of self-supervised learning and information theory. They concluded that by scaling the capacity of ViTs the out-of-distribution (OOD) generalization performance can be enhanced, mostly under the domain adaptation settings. 
On the other hand, we show that it is possible to improve the OOD generalization performance of ViTs without introducing any new parameters under the established DG protocols \cite{Gulrajani2021InSO}. 

\noindent\textbf{Knowledge Distillation:} was initially designed for model compression and aims at matching the output of a teacher model to a student model whose capacity is smaller than the teacher model \cite{hinton2015distilling}.
Zhang \etal. \cite{zhang2019your} partitioned a CNN model into several blocks, and the knowledge from the full (deeper part) of the model is squeezed into the shallow parts.
Yun \etal. \cite{yun2020regularizing} proposed a self-distillation approach based on penalizing the predictive distributions between similar data samples. In particular, it distills the predictive distribution between different samples of the same label during training.
Towards addressing DG problem, Wang \etal. \cite{wang2021embracing} proposed a teacher-student distillation strategy, based on CNNs, and a gradient filter as an efficient regularization term.
In contrast, we propose a new self-distillation strategy to enhance the DG capabilities of ViTs. It prevents introducing any new parameters via seamlessly exploiting the modular architecture of ViTs.

\section{Proposed Approach} \label{proposed_method}
In this paper, we aim to explore ViTs towards tackling the domain generalization problem. We observe that a simple, but competitive DG baseline (ERM) built on ViT displays notable performance decay in a typical DG setting (Fig.~\ref{fig:in_domain_out_domain_comparison}). Towards this end, we propose a simple plug-and-play DG approach for ERM-ViT, termed as self-distillation for ViTs, that explicitly facilitates the model towards exploiting cross-domain transferable features (Fig.~\ref{fig:motivation_fig}).

\subsection{Preliminaries}
\label{subsection:Preliminaries}

\textbf{Problem Settings:} In traditional domain generalization (DG) setting \cite{Gulrajani2021InSO}, we assume the availability of data from a set of training (source) domains $\mathcal{D}=\{\mathcal{D}\}_{k=1}^{K}$. Where $\mathcal{D}_{k}$ denotes a distribution over the input space $\mathcal{X}$ and $K$ is the total number of training domains. From a domain $k$, we sample $J$ training datapoints that comprise of input $x$ and label $y$ as pairs $(x_{j}^{k} \in \mathcal{X},y_{j}^{k}\in \mathcal{Y})_{j=1}^{J}$. Besides a set of training (source) domains, we also assume a set of target domains $\{\mathcal{T}\}_{t=1}^{T}$, where $T$ is the total number of target domains and is typically set to 1. The goal in DG is to learn a mapping ${\mathcal{F}}_{\theta}: \mathcal{X} \rightarrow \mathcal{Y}$ that provides accurate predictions on data from an unseen target domain $\mathcal{T}_{t}$.

\noindent \textbf{Empirical risk minimization (ERM) for DG:} Assume a loss function $\mathcal{L}:\mathcal{Y}\times\mathcal{Y}$ which can quantify the prediction error, such as standard Cross Entropy (CE) for the image recognition task. A simple DG baseline accumulates the data from multiple source domains $\mathcal{D}$ and searches for a predictor minimizing the following empirical risk \cite{vapnik1999nature}: $ \frac{1}{N}\sum_{i=1}^{N}\mathcal{L}(\mathcal{F}_{\theta}(x_{k}^{j},y_{k}^{j}))$. Where $N=K \times J$ is the total number of data points from all source domains. Recently, Gulrajani and Lopez-Paz \cite{Gulrajani2021InSO} demonstrated that this simple ERM based DG baseline shows competitive results or even performs better than many previous state-of-the-art DG methods under a fair evaluation protocol.

\noindent \textbf{ViT based ERM:} While exploring ViTs for DG, we observe that ViT-based ERM (ERM-ViT) shows notable performance drop, similar to their CNN-based ERM counterpart (Fig.~\ref{fig:in_domain_out_domain_comparison}). This is likely due to the lack of any explicit overfitting mechanism and the supervision from one-hot encoded ground truth labels. It renders the overall learning of the mapping problem, from input space to label space, rather difficult. As a result, the model is more prone to exploiting non-generalizable, brittle features, such as the specific background of a domain (Fig.~\ref{fig:motivation_fig} \&~\ref{fig:attention_maps}). To tackle this problem, in the next section, we propose a new self-distillation technique for improving the DG capabilities of ViTs. The core idea is to ease the mapping problem by generating non-zero entropy supervision for multiple feature pathways in ViTs. This enables the model towards utilizing more generalizable features (Fig.~\ref{fig:motivation_fig} \&~\ref{fig:attention_maps}), that are mostly shared across source and target domains.

\subsection{Self-distilled Vision Transformer for Domain Generalization}
\label{subsection:Self-distilled Vision Transformers for Domain Generalization}

\noindent \textbf{ViTs have modular architectures:} Assume the model $\mathcal{F}$ is composed of $n$ intermediate layers and the final classifier $h$ such as $\mathcal{F} = (f_{1} \circ f_{2} \circ f_{3} \circ \ldots f_{n}) \circ h$, where $f_{i}$ represents an intermediate block or layer. In the case of ViT (\eg ~Deit-Small \cite{touvron2021training}), $f_{i}$ is based on a self-attention transformer block, and such network design is monolithic as any transformer block produces equi-dimensional features that are $\mathbb{R}^{m \times d}$, where $m$ represent the number of input features or tokens  and each token has $d$ dimensions. The monolithic design approach of ViT allows a self-ensemble behavior \cite{naseer2021improving}, where the output of each block can be processed by the final classifier $h$ to create an intermediate classifier\footnote{For non-monolithic ViT designs and CNNs, where feature dimension changes across the layers, the intermediate classifier can be obtained via $\mathcal{F}_{i} = (f_{i} \circ g_{i}) \circ h$, where $g$ projects the output of $f_{i}$ to the same dimension as $h$.}.
\begin{equation} \label{eq: sub-model}
    \mathcal{F}_{i} = f_{i} \circ h
\end{equation}
Our goal is to induce the so-called dark knowledge, non-zero entropy supervisory signals, from the final classifier to these sub-models, manifesting multiple feature pathways.

\noindent\textbf{Self-distillation in ViTs:} 
As discussed earlier, ViTs can be easily dissected into a number of sub-models due to their monolithic architectural design as each transformer block produces a classification token that can be processed by the final classification head (Eqn. \ref{eq: sub-model}) to produce a class-specific score.
Each sub-model represents a discriminative feature pathway through the network. We believe that inducing dark knowledge from the final classifier output to these sub-models via soft supervision during training can enhance the overall model's capability towards learning object semantics.

\begin{figure}[t] 
  \centering
  \includegraphics[width=\textwidth]{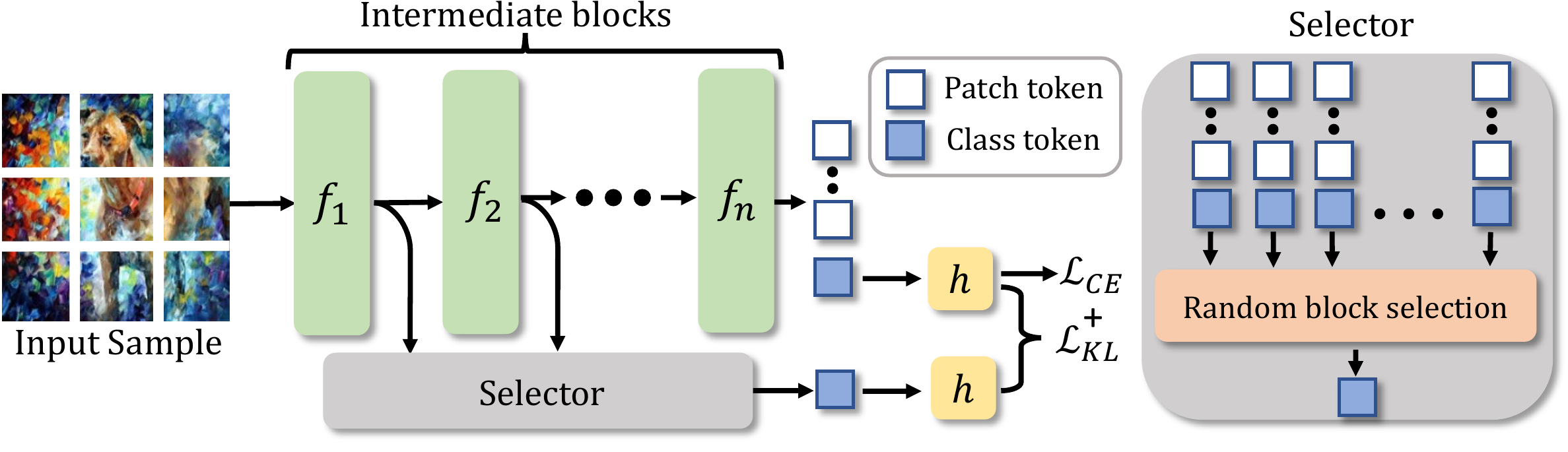}
  \caption{\small Proposed self-distillation in ViTs for domain generalization (ERM-SDViT). ViTs build upon a modular and hierarchical architecture, where a model is comprised of $n$ intermediate blocks/layers ($f_{i}$) and a final classifier $h$.  
  The 'Selector' chooses a random block from the range of intermediate blocks and makes a prediction after passing its classification token through the final classifier. This way the dark knowledge, as non-zero entropy signals, is distilled from the final classification token to the intermediate class tokens during training.} 
  \label{model_figure}
\end{figure}

\noindent\textbf{Random sub-model distillation:} The number of sub-models within a given ViT depends on the number of Transformer blocks (see Fig.~\ref{model_figure}) and distilling the knowledge to all of the sub-models at once poses optimization difficulties during online training.  
Therefore, we introduce a simple technique that randomly samples one sub-model based on Eqn. \ref{eq: sub-model} from all the possible set of sub-models (see Fig.~ \ref{model_figure}). In this manner, our approach trains all sub-models but knowledge is transferred to only a single sub-model at any step of the training. This strategy eases the optimization and leads to better domain generalization.

\begin{figure}[t]
  \centering
  \includegraphics[width=\textwidth]{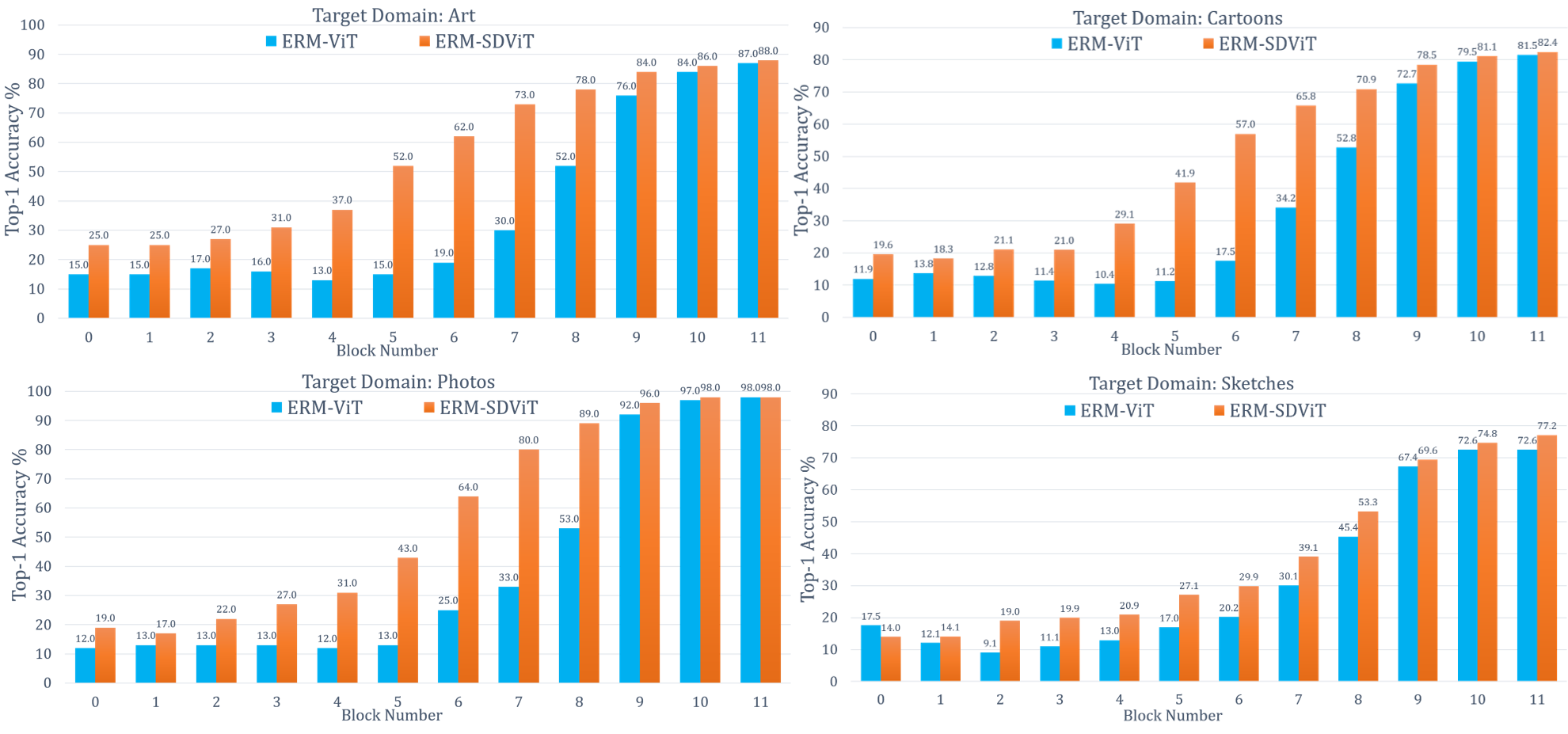}
  \caption{\small Block-wise accuracy of baseline (ERM-ViT \cite{touvron2021training}) and our method (ERM-SDViT), featuring random sub-model distillation for DG. Results are reported on four challenging target domains from \texttt{PACS} dataset.}
  \label{fig:Blk_accu}
  \end{figure}

\noindent \textbf{Impact on internal representations:} In Fig.~\ref{fig:Blk_accu}, we plot the block-wise accuracy of baseline (ERM-ViT) and our method (ERM-SDViT). Random sub-model distillation improves the accuracy of all blocks, in particular, the improvement is more pronounced for the earlier blocks. Besides later blocks, it also encourages earlier blocks to bank on transferable representations, yet discriminative representations. Since these earlier blocks manifest multiple discriminative feature pathways, we believe that they better facilitate the overall model towards capturing the semantics of the object class.

\noindent\textbf{Training Objective:} For a given input $x$, the prediction error from the final classification token of the ViT is computed  using cross-entropy loss in comparison with one-hot encoded ground-truths as follows.
\begin{equation}\label{eq: ce}
\mathcal{L}_{\mathrm{CE}}(\mathcal{F}(x), y) = - \sum_{j=1}^n y_jlog(\mathcal{F}(x)_j),
\end{equation}
where $n$ is the output dimension of the final classifier. We randomly sample an intermediate block as shown in Fig.~\ref{model_figure} and produce logits from  the classification token of a sub-model by applying Eqn.\ref{eq: sub-model}.

We then compute the difference between the final and randomly sampled intermediate classification token by comparing the KL divergence between their logit distributions as follows:

\begin{equation} 
\label{eq:kl}
    \mathcal{L}_{\mathrm{KL}}(\mathcal{F}(x)\| \mathcal{F}_i(x))  =  \sum\limits_{j=1}^n \sigma\left( \mathcal{F}(x)/\tau \right)_j \log\frac{\sigma\left(\mathcal{F}(x)/\tau\right)_{j}}{\sigma\left(\mathcal{F}_i(x)/\tau\right)_{j}}, 
\end{equation} 
where $\sigma$ denotes the softmax operation and $\tau$ represent temperature used to rescale the logits \cite{hinton2015distilling}. The model is optimized by minimizing the overall loss  based on Eqns. \ref{eq: ce} and \ref{eq:kl} and given as follows:
\begin{equation}
    \mathcal{L} = \mathcal{L}_{\mathrm{CE}} + \lambda \mathcal{L}_{\mathrm{KL}},
\end{equation}
where $\lambda$ balances the contribution of $\mathcal{L}_{\mathrm{KL}}$ towards the overall loss $\mathcal{L}$.

\section{Experiments}
\label{section:Experiments}

\noindent \textbf{Datasets:} Following the work of Gulrajani and Lopez-Paz \cite{Gulrajani2021InSO}, we rigorously evaluate the effectiveness of our proposed method and draw comparisons with the existing state-of-the-art on five benchmark datasets including \texttt{PACS} \cite{li2017deeper}, \texttt{VLCS} \cite{fang2013unbiased}, \texttt{OfficeHome} \cite{venkateswara2017deep}, \texttt{TerraIncognita} \cite{beery2018recognition} and \texttt{DomainNet} \cite{peng2019moment}.
\texttt{PACS} \cite{li2017deeper} contains four domains $d \in$ \{Art, Cartoons, Photos, Sketches\},7 classes and a total of 9,991 images. \texttt{VLCS} \cite{fang2013unbiased} comprises of four domains as well $d \in$ \{Caltech101, LabelMe, SUN09, VOC2007\}, 5 classes and offers 10,729 images. \texttt{OfficeHome} \cite{venkateswara2017deep} also contains four domains $d \in$ \{Art, Clipart, Product, Real\}, 65 classes and a total of 15,588 images. 
\texttt{TerraIncognita} \cite{beery2018recognition}: has four camera-trap domains $d \in$ \{L100, L38, L43, L46\}, 10 classes and offers 24,778 wild photographs. \texttt{DomainNet} \cite{peng2019moment} contains six domains $d \in$ \{Clipart, Infograph, Painting, Quickdraw, Real, Sketch\}, 345 classes and 586,575 images.

\noindent \textbf{Implementation and training/testing details:} To allow fair comparisons, we follow the training and evaluation protocol of Gulrajani and Lopez-Paz \cite{Gulrajani2021InSO}.
We use the training domain validation protocol for model selection. After partitioning each training domain data into the training and validation subsets (80\%/20\%), the validation data from each training domain are pooled to obtain an overall validation set. The model that maximizes the accuracy on this overall validation set is considered the best model which is then evaluated on the target domain to report classification (top-1) accuracy.
For all our ViT-based methods, including the proposed approach, we use AdamW \cite{loshchilov2018fixing} optimizer and use the default hyperparameters (HPs) of ERM from \cite{Gulrajani2021InSO}
\footnote{They are default parameters in the pre-defined ranges \cite{Gulrajani2021InSO} for random HP search.}, including the batch size of $32$, the learning rate of $5e$-$05$, and the weight decay of $0.0$. Note that, only the values of our method-specific HPs, $\lambda$ and $\tau$, are sought via grid search in the ranges $\{0.1, 0.2, 0.5 \}$ and $\{ 3.0, 5.0\}$, respectively, using the validation set.
We report accuracy for each target domain and their average where a model is trained/validated on training domains and evaluated on an (unseen) target domain. Each accuracy on the target domain is an average over three different trials with different train-validation splits.

\noindent \textbf{Evaluation with different ViT backbones:}
We establish the generalizability of our method by experimenting with three different ViT backbones, namely \textbf{DeiT} \cite{touvron2021training}, \textbf{CvT} \cite{wu2021cvt}, and \textbf{T2T-ViT} \cite{yuan2021tokens}. \textbf{DeiT} is a data-efficient image transformer and was trained on 1.2 million ImageNet examples. We use the DeiT-Small model having 22M parameters, which can be regarded as a ResNet-50 counterpart containing 23.5M parameters. Note that, we utilize the DeiT-Small model without the distillation token and a student-teacher formulation.

\textbf{CvT} introduces convolutions into ViT to improve accuracy and efficiency. We use CvT-21 which contains 32M parameters in our baselines and proposed self-distilled ViTs.
\textbf{T2T-ViT} relies on a progressive tokenization to aggregate neighboring Tokens to one Token; it can encode the local structure information of surrounding tokens and reduce the length of tokens iteratively. We use T2T-ViT-14 model, containing 21.5M parameters, which is approx. equivalent to the capacity of the ResNet-50 model.

\begin{table}[t]
    \centering
        \caption{\small Comparison with the several (17) existing SOTA DG methods. The best results are in bold and the second best is underlined.}
    \label{SOTA_comparison_table}
    \small
    \tabcolsep=0.05cm
    \adjustbox{max width=\textwidth}{
    \begin{tabular}{lp{0.1cm}cp{0.1cm}cp{0.1cm}cp{0.1cm}cp{0.1cm}cp{0.1cm}cp{0.1cm}cp{0.1cm}c}
\toprule
\rowcolor{Gray} 
Algorithm           && Backbone            && \# Params            && \texttt{VLCS}    && \texttt{PACS}             && \texttt{OfficeHome}       && \texttt{TerraInc}   && \texttt{DomainNet}        && Average              \\
\midrule
ERM \cite{Gulrajani2021InSO}          && ResNet-50 &&  23.5M    && 77.5 $\pm$ 0.4            && 85.5 $\pm$ 0.2            && 66.5 $\pm$ 0.3            && 46.1 $\pm$ 1.8            && 40.9 $\pm$ 0.1            && 63.3                      \\
IRM \cite{arjovsky2019invariant}               && ResNet-50 &&  23.5M     && 78.5 $\pm$ 0.5            && 83.5 $\pm$ 0.8            && 64.3 $\pm$ 2.2            && 47.6 $\pm$ 0.8            && 33.9 $\pm$ 2.8            && 61.5                       \\
GroupDRO \cite{sagawa2019distributionally}          && ResNet-50 &&  23.5M       && 76.7 $\pm$ 0.6            && 84.4 $\pm$ 0.8            && 66.0 $\pm$ 0.7            && 43.2 $\pm$ 1.1            && 33.3 $\pm$ 0.2            &&   60.7                   \\
Mixup \cite{yan2020improve}              && ResNet-50 &&  23.5M      && 77.4 $\pm$ 0.6            && 84.6 $\pm$ 0.6            && 68.1 $\pm$ 0.3            && 47.9 $\pm$ 0.8            && 39.2 $\pm$ 0.1            &&  63.4                     \\
MLDG \cite{li2018learning}              && ResNet-50 &&  23.5M       && 77.2 $\pm$ 0.4            && 84.9 $\pm$ 1.0            && 66.8 $\pm$ 0.6            && 47.7 $\pm$ 0.9            && 41.2 $\pm$ 0.1            &&       63.5                \\
CORAL \cite{sun2016deep}            && ResNet-50 &&  23.5M        && 78.8 $\pm$ 0.6            && 86.2 $\pm$ 0.3            && 68.7 $\pm$ 0.3            && 47.6 $\pm$ 1.0            && 41.5 $\pm$ 0.1            &&   64.5                    \\
MMD \cite{li2018domain}             && ResNet-50 &&  23.5M         && 77.5 $\pm$ 0.9            && 84.6 $\pm$ 0.5            && 66.3 $\pm$ 0.1            && 42.2 $\pm$ 1.6            && 23.4 $\pm$ 9.5            && 58.8                     \\
DANN \cite{ganin2016domain}          && ResNet-50 &&  23.5M           && 78.6 $\pm$ 0.4            && 83.6 $\pm$ 0.4            && 65.9 $\pm$ 0.6            && 46.7 $\pm$ 0.5            && 38.3 $\pm$ 0.1            &&     62.6                  \\
CDANN \cite{li2018deep}          && ResNet-50 &&  23.5M           && 77.5 $\pm$ 0.1            && 82.6 $\pm$ 0.9            && 65.8 $\pm$ 1.3            && 45.8 $\pm$ 1.6            && 38.3 $\pm$ 0.3            &&       62.0                \\
MTL \cite{blanchard2017domain}              && ResNet-50 &&  23.5M         && 77.2 $\pm$ 0.4            && 84.6 $\pm$ 0.5            && 66.4 $\pm$ 0.5            && 45.6 $\pm$ 1.2            && 40.6 $\pm$ 0.1            &&    62.8                   \\
SagNet \cite{Nam_2021_CVPR}   && ResNet-50 &&  23.5M                && 77.8 $\pm$ 0.5            && 86.3 $\pm$ 0.2            && 68.1 $\pm$ 0.1            && 48.6 $\pm$ 1.0            && 40.3 $\pm$ 0.1            &&        64.2               \\
ARM \cite{zhang2021adaptive}              && ResNet-50 &&  23.5M        && 77.6 $\pm$ 0.3            && 85.1 $\pm$ 0.4            && 64.8 $\pm$ 0.3            && 45.5 $\pm$ 0.3            && 35.5 $\pm$ 0.2            &&  61.7                     \\
VREx \cite{krueger2021out}             && ResNet-50 &&  23.5M         && 78.3 $\pm$ 0.2            && 84.9 $\pm$ 0.6            && 66.4 $\pm$ 0.6            && 46.4 $\pm$ 0.6            && 33.6 $\pm$ 2.9            &&        61.9               \\
RSC \cite{huang2020self}           && ResNet-50 &&  23.5M           && 77.1 $\pm$ 0.5            && 85.2 $\pm$ 0.9            && 65.5 $\pm$ 0.9            && 46.6 $\pm$ 1.0            && 38.9 $\pm$ 0.5            &&     62.6                  \\
SelfReg \cite{Kim_2021_ICCV} && ResNet-50 && 23.5M && 77.5 $\pm$ 0.0 && 86.5 $\pm$ 0.3&& 69.4 $\pm$ 0.2 && \underline{51.0 $\pm$ 0.4} && 44.6 $\pm$ 0.1 && 65.8\\
mDSDI \cite{bui2021exploiting} && ResNet-50 && 23.5M && 79.0 $\pm$ 0.3 &&  86.2 $\pm$ 0.2 && 69.2 $\pm$ 0.4 && 48.1 $\pm$ 1.4 && 42.8 $\pm$ 0.1 && 65.0\\ 
SWAD \cite{cha2021swad} && ResNet-50 && 23.5M && 79.1 $\pm$ 0.1 && 88.1 $\pm$ 0.1 && 70.6 $\pm$ 0.2 && 50.0 $\pm$ 0.3 && 46.5 $\pm$ 0.1 && 66.8\\ \hline
ERM-ViT \cite{touvron2021training} && DeiT-Small && 22M && 78.8 $\pm$ 0.5 && 84.9 $\pm$ 0.9 && 71.4 $\pm$ 0.1 &&  43.4 $\pm$ 0.5 && 45.5 $\pm$ 0.0&& 64.8\\
ERM-ViT + T3A && DeiT-Small && 22M && 81.6 $\pm$ 0.2 && 85.5 $\pm$ 0.7 && 72.6 $\pm$ 0.2 && 43.6 $\pm$ 0.4 && 46.8 $\pm$ 0.1 && 66.0\\
ERM-SDViT && DeiT-Small && 22M && 78.9 $\pm$ 0.4  &&  86.3 $\pm$ 0.2   &&  71.5 $\pm$ 0.2 && 44.3 $\pm$ 1.0 && 45.8 $\pm$ 0.0 && 65.3 \\
ERM-SDViT + T3A && DeiT-Small && 22M && \underline{81.6 $\pm$ 0.1}  && 86.7 $\pm$ 0.2 && 72.5 $\pm$ 0.3 && 44.9 $\pm$ 0.4 && 47.4 $\pm$ 0.1 && 66.6\\
ERM-ViT \cite{wu2021cvt} && CvT-21 && 32M && 79.0 $\pm$ 0.3  && 86.9 $\pm$ 0.3 && 75.5 $\pm$ 0.0 && 48.7 $\pm$ 0.4&& 50.4 $\pm$ 0.1 && 68.1 \\
ERM-ViT + T3A && CvT-21 && 32M && 80.6 $\pm$ 0.3 && \underline{88.5 $\pm$ 0.1} && \underline{76.2 $\pm$ 0.0} && 49.7 $\pm$ 0.5 && 52.0 $\pm$ 0.1  && \underline{69.4} \\
ERM-SDViT && CvT-21 && 32M && 79.2 $\pm$ 0.4  && 88.3 $\pm$ 0.2 &&  75.6 $\pm$ 0.2  &&  49.7 $\pm$ 1.4 && \underline{50.4 $\pm$ 0.0 } && 68.6 \\
ERM-SDViT + T3A && CvT-21 && 22M && \textbf{81.9 $\pm$ 0.4} && \textbf{88.9 $\pm$ 0.5} && \textbf{77.0 $\pm$ 0.2} &&\textbf{51.4 $\pm$ 0.7} && \textbf{52.0 $\pm$ 0.0} && \textbf{70.2} \\
ERM-ViT \cite{yuan2021tokens} && T2T-ViT-14 && 21.5M && 78.9 $\pm$ 0.3 && 86.8 $\pm$ 0.4  &&  73.7 $\pm$ 0.2 && 48.1 $\pm$ 0.2 && 48.1 $\pm$ 0.1 && 67.1\\
ERM-ViT + T3A && T2T-ViT-14 && 21.5M && 81.0 $\pm$ 0.6 && 87.7 $\pm$ 0.4 && 75.3 $\pm$ 0.1 && 47.8 $\pm$ 0.2 && 50.0 $\pm$ 0.1  && 68.3\\
ERM-SDViT && T2T-ViT-14 && 21.5M && 79.5 $\pm$ 0.8   &&  88.0 $\pm$ 0.7 &&   74.2 $\pm$  0.3   && \underline{50.6 $\pm$ 0.8} &&   48.2 $\pm$ 0.2 && 68.1 \\
ERM-SDViT + T3A && T2T-ViT-14 && 21.5M && 81.2 $\pm$ 0.3  && 87.8 $\pm$ 0.6 && 75.5 $\pm$ 0.2  && 50.5 $\pm$ 0.6&& 50.2 $\pm$ 0.1 && 69.0\\
\bottomrule

\end{tabular}}
\end{table}

\subsection{Comparison with the state-of-the-art}
\label{subsection:Comparison with the state-of-the-art}
We compare our approach to several (in particular, 17) existing state-of-the-art algorithms for DG (see Table~\ref{SOTA_comparison_table}) listed in Domainbed suite \cite{Gulrajani2021InSO}. 
Specifically, we include the following DG algorithms: Empirical Risk Minimization (ERM) \cite{Gulrajani2021InSO}, Invariant Risk Minimization (IRM) \cite{arjovsky2019invariant}, Group Distributionally Robust Optimization (GroupDRO) \cite{sagawa2019distributionally}, Inter-domain Mixup (Mixup) \cite{yan2020improve},
Meta-Learning for Domain Generalization (MLDG) \cite{li2018learning}, Deep CORrelation ALignment (CORAL) \cite{sun2016deep}, Maximum Mean Discrepancy (MMD) \cite{li2018domain}, Domain Adversarial Neural Networks (DANN) \cite{ganin2016domain}, Class-conditional DANN (CDANN) \cite{li2018deep}, Marginal Transfer Learning (MTL) \cite{blanchard2017domain}, Style-Agnostic Networks (SagNet) \cite{Nam_2021_CVPR}, Adaptive Risk Minimization (ARM) \cite{zhang2021adaptive}, Variance Risk Extrapolation (VREx) \cite{krueger2021out}, Representation Self Challenging (RSC) \cite{huang2020self}, Self-supervised
contrastive regularization (SelfReg) \cite{Kim_2021_ICCV}, meta-Domain Specific-Domain Invariant (mDSDI) \cite{bui2021exploiting}, and Stochastic Weight Averaging Densely (SWAD) \cite{cha2021swad}.

\noindent \textbf{\texttt{VLCS and} \texttt{OfficeHome}:} In \texttt{VLCS}, our approach (ERM-SDViT) records the best classification accuracy of 79.5\% with T2T-ViT-14 backbone outperforming the baseline (ERM-ViT) and DG SOTA algorithms. Similarly, in \texttt{OfficeHome}, our method outperforms all other methods under all three ViT backbones. In particular, it displays the best accuracy of 75.6\% with CvT-21 backbone.

\noindent \textbf{\texttt{PACS,} \texttt{DomainNet and} \texttt{TerraInc}:} In \texttt{PACS} our approach delivers the top accuracy of 88.3\% and in \texttt{DomainNet} it achieves an accuracy of 50.4\% with CvT-21 backbone. In \texttt{TerraIncognita}, our method achieves a competitive accuracy of 50.6\% with T2T-ViT-14 backbone against the top performing method of SelfReg \cite{Kim_2021_ICCV}. In the overall average accuracy over five datasets, our method outperforms the existing state-of-the-art in DG with CvT-21 and T2T-ViT-14 backbones. Moreover, it provides notable gains over the baseline (ERM-ViT) under the three recent ViT backbone architectures.  

\begin{figure}[t]
  \centering
  \includegraphics[width=\textwidth]{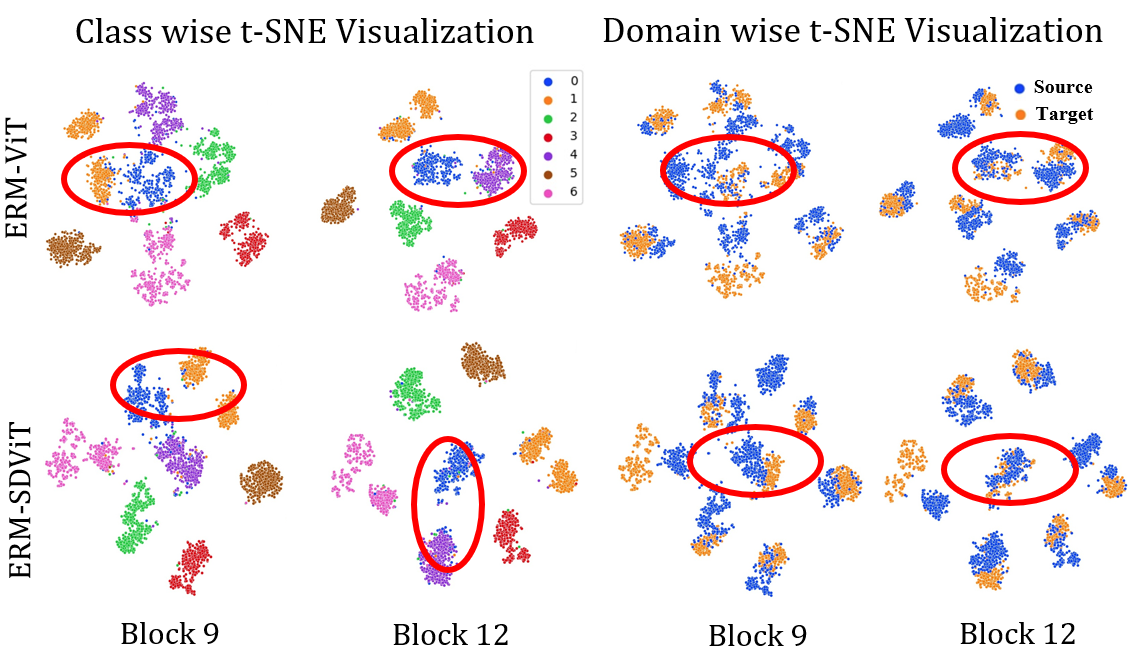}
  \caption{\small t-SNE visualization of features from different blocks (9 \& 12) in baseline and our approach. Left: Features are colored corresponding to their class labels (classes: 7,\texttt{PACS} dataset). Right: Features are colored corresponding to their domain labels. Our approach has performed well for instance, in class-wise t-SNE in block 9, the features of class 0 and 1 (highlighted in red circle) are well separated as compared to ERM-ViT baseline. Similarly in class 0 and 4 in the final $12^{th}$ block features of our ERM-SDViT approach are also separated clearly. While in domain-wise t-SNE, a similar pattern is observed, as source and target domain features are more overlapped with each other and clearly separated as well. See Appendix for more t-SNE results.}
  \label{fig:Tsne_cls}
\end{figure}

\begin{figure}[t]
  \centering
  \includegraphics[width=\textwidth]{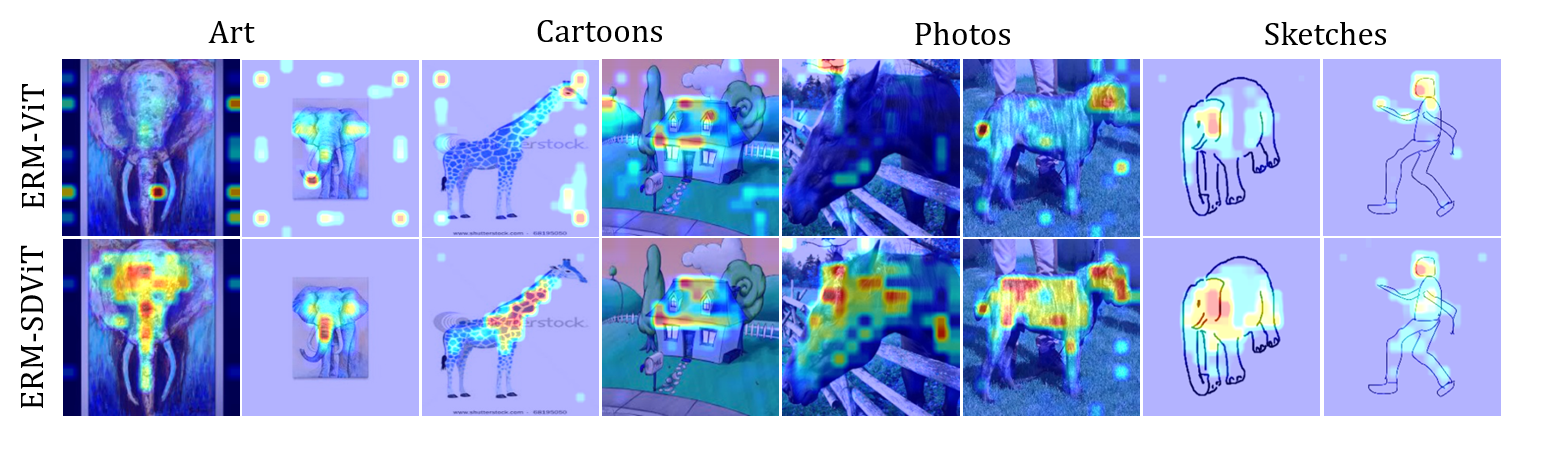}
  \caption{\small Attention maps from baseline (ERM-ViT) and proposed (ERM-SDViT, backbone: DeiT-Small). They are computed at the final block of ViT.} 
  \label{fig:attention_maps}
\end{figure}  \label{fig:attention_maps}
\begin{figure}[t]
  \centering
  \includegraphics[width=\textwidth]{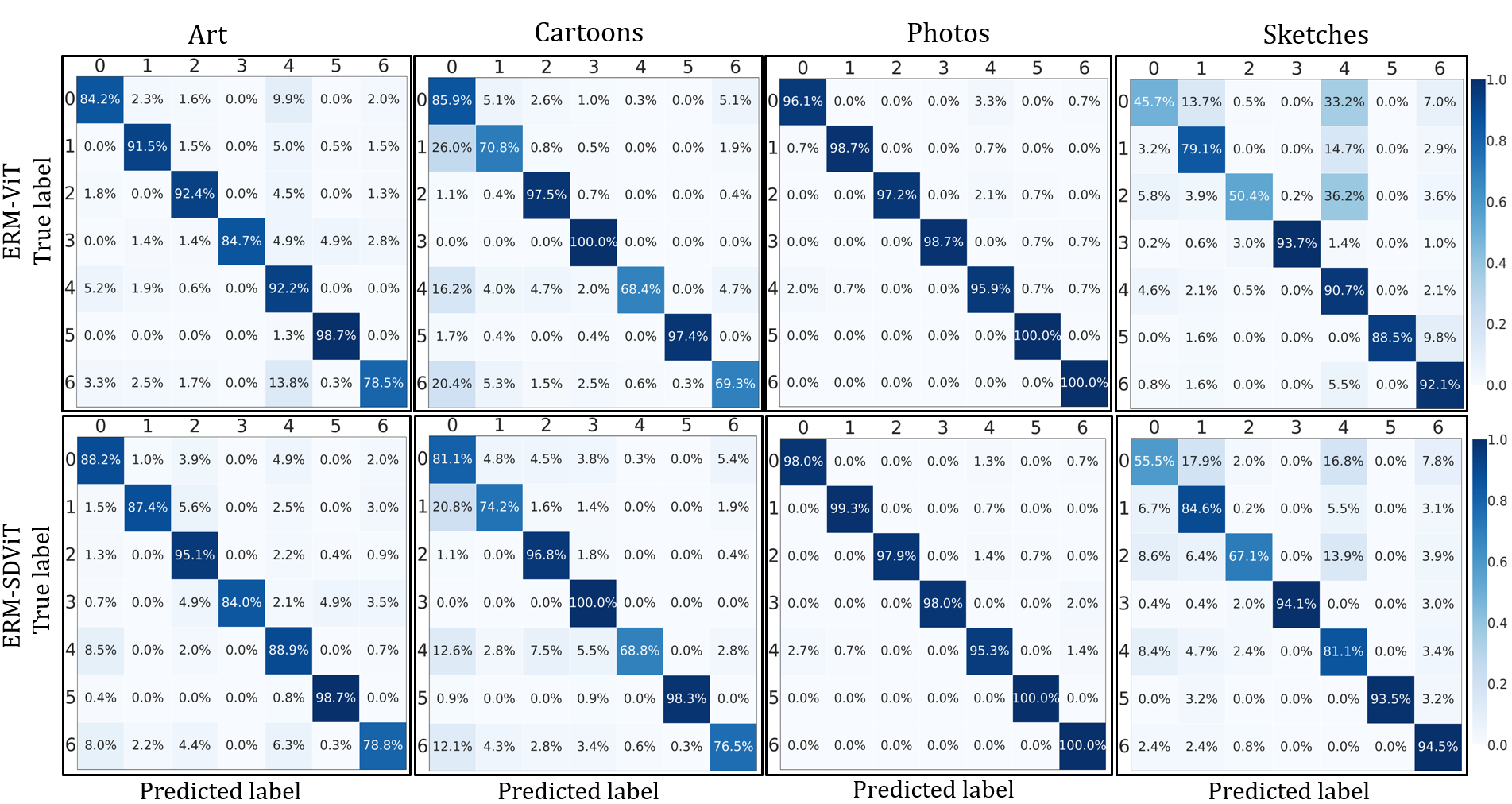}
  \caption{\small Confusion matrices for baseline and our method. The classes in the Figure are `0':Dog, `1': Elephant, `2':Giraffe, `3':Guitar, `4':Horse, `5':House, and `6':Person.}  \label{fig:confusion_matrix_PACS_average}
\end{figure}
\subsection{Ablation Study and Analysis}
\label{subsection:Ablation_Study} 
In all experiments, unless stated otherwise, the baseline method is ERM-ViT with DeiT-Small backbone \cite{touvron2021training}. 

\begin{table}[t]
\begin{center}
\caption{\small Our proposed approach is also effective in further
improving the performance of a strong DG baseline namely T3A \cite{iwasawa2021test}. Results are reported with different ERM-ViT backbone architectures: DeiT-Small, CvT-21, and T2T-ViT-14.}
\label{tab:ablation_PACS}
\adjustbox{max width=\textwidth}{
\begin{tabular}{lccccccc}
\hline \noalign{\smallskip}
\rowcolor{Gray}
Model & Backbone & \# of Params & Art & Cartoons & Photos & Sketches & Average \\
\noalign{\smallskip}
\hline
\noalign{\smallskip}
ERM  & ResNet-50 & 23.5M & 81.3 $\pm$ 0.6 & 80.9 $\pm$ 0.3& 96.3 $\pm$ 0.6  &  78.0 $\pm$ 1.6 & 84.1 $\pm$ 0.4  \\
ERM-ViT  & DeiT-Small & 22M &  87.4 $\pm$ 1.2 &81.5 $\pm$ 0.8&	98.1 $\pm$ 0.1	& 72.6 $\pm$ 3.3&	84.9 $\pm$ 0.9   \\
ERM-ViT + T3A & DeiT-Small & 22M & 88.1 $\pm$ 1.5 & 81.8 $\pm$ 0.8 & 98.3 $\pm$ 0.1 & 73.8 $\pm$ 2.7 & 85.5 $\pm$ 0.7 \\
ERM-SDViT & DeiT-Small & 22M & 87.6 $\pm$ 0.3   &       82.4 $\pm$ 0.4      &    98.0 $\pm$ 0.3     &     77.2 $\pm$ 1.0 & 86.3 $\pm$ 0.2     \\
ERM-SDViT + T3A & DeiT-Small & 22M &  88.2 $\pm$ 0.4   &       82.9 $\pm$ 0.5    &      98.3 $\pm$ 0.1     &     77.2 $\pm$ 0.9    &      86.7 $\pm$ 0.2 \\
ERM-ViT  & CvT-21 & 32M &  89.0 $\pm$ 1.0   &       84.8 $\pm$ 0.6   &       98.8 $\pm$ 0.2   &       78.6 $\pm$ 0.3   &   87.8 $\pm$ 0.1  \\
ERM-ViT  + T3A & CvT-21 & 32M & 90.1 $\pm$ 0.7  &         85.3 $\pm$ 0.6   &    99.0 $\pm$ 0.1  & 79.6 $\pm$ 0.4 & \underline{88.5 $\pm$ 0.1} \\
ERM-SDViT & CvT-21 & 32M &  90.8 $\pm$ 0.1   &       84.1 $\pm$ 0.5    &      98.3 $\pm$ 0.2      &    80.0 $\pm$ 1.3    &   88.3 $\pm$ 0.2   \\
ERM-SDViT + T3A & CvT-21 & 32M  & 91.2 $\pm$ 0.8   &       83.5 $\pm$ 0.2     &     98.3 $\pm$ 0.1    &      82.5 $\pm$ 1.5   & \textbf{88.9 $\pm$ 0.5}     \\
ERM-ViT  & T2T-ViT-14 &  21.5M & 89.6 $\pm$ 0.9 & 81.0 $\pm$ 0.9 &  98.9 $\pm$ 0.2    &  77.6 $\pm$ 2.6   & 86.8 $\pm$ 0.4    \\
ERM-ViT  + T3A & T2T-ViT-14 & 21.5M & 90.7 $\pm$ 1.0   & 82.4 $\pm$ 0.6  &  99.1 $\pm$ 0.1  &    78.5 $\pm$ 2.2    &  87.7 $\pm$ 0.4   \\
ERM-SDViT & T2T-ViT-14 & 21.5M & 90.2 $\pm$ 1.2  &        82.7 $\pm$ 0.7   &       98.6 $\pm$ 0.2       &   80.5 $\pm$ 2.2    &      88.0  $\pm$ 0.7  \\
ERM-SDViT + T3A & T2T-ViT-14 & 21.5M & 89.2 $\pm$ 1.8  & 84.0 $\pm$ 0.1 &   98.7 $\pm$ 0.1   &       79.3 $\pm$ 1.1 & 87.8 $\pm$ 0.6   \\

\hline
\end{tabular}
}
\end{center}
\end{table}

\begin{table}[t]
\begin{center}
\caption{\small Performance with different block selection strategies in our self-distillation.}
\label{ablation_Distillation}
\adjustbox{max width=\textwidth}{
\begin{tabular}{lccccc}
\hline \noalign{\smallskip}
\rowcolor{Gray}
Model  & Art & Cartoons & Photos & Sketches & Average \\
\hline
ERM-ViT &  87.4 $\pm$ 1.2 &81.5 $\pm$ 0.8&	98.1 $\pm$ 0.1	& 72.6 $\pm$ 3.3&	84.9 $\pm$ 0.9   \\
ERM-SDViT[0-5] &  87.3 $\pm$ 0.2 &  82.1 $\pm$ 0.4   &       98.3 $\pm$ 0.1      &    76.6 $\pm$ 2.1  &  \underline{86.1 $\pm$ 0.5}   \\
ERM-SDViT[6-11]  & 86.8 $\pm$ 0.8  &   81.4 $\pm$ 0.3  &        98.6 $\pm$ 0.1     &     74.3 $\pm$ 1.7 &  85.3 $\pm$ 0.3      \\
ERM-ViT[self.dist all blocks] & 87.8 $\pm$ 1.9    &      82.2 $\pm$ 0.7    &      97.9 $\pm$ 0.1    &      75.0 $\pm$ 1.1    &      85.7  $\pm$ 0.3   \\
ERM-SDViT[0-11] Ours & 87.6 $\pm$ 0.3   &       82.4 $\pm$ 0.4      &    98.0 $\pm$ 0.3     &     77.2 $\pm$ 1.0 & \textbf{86.3 $\pm$ 0.2}     \\
\noalign{\smallskip}
\hline
\end{tabular}
}
\end{center}
\end{table}

\noindent \textbf{With a recent DG baseline:} We show that our proposed approach is also effective in further improving the performance of a strong DG baseline namely T3A \cite{iwasawa2021test} (see Table~\ref{SOTA_comparison_table} and ~\ref{tab:ablation_PACS}). T3A computes a pseudo-prototype representation for each class using unlabeled data augmented by the base classifier trained in the source domains in an online manner. A test image is classified based on its distance to the pseudo-prototype representation. Our proposed approach with T3A (ERM-SDViT+T3A) consistently improves the performance of the baseline (ERM-ViT+T3A) with all three ViT backbones.

\noindent \textbf{Feature visualizations:} Fig.~\ref{fig:Tsne_cls} (left) visualizes the class-wise feature representations of different blocks using t-SNE in baseline and our approach. Our approach facilitates learning more discriminative features. As a result, the intra-class features are compactly clustered while the inter-class features are far apart.
Likewise, fig. \ref{fig:Tsne_cls} (right) visualizes the same features, however, they are colored based on their domain labels. Our method promotes a greater overlap between the features of source and target domains. Moreover, we quantify the domain overlap between the source/target features using cosine similarity (see Tab.~\ref{tab:cosine_table1}).

\begin{SCtable}
\caption{\small Domain overlap quantified based on cosine similarity between source and target domain class tokens.}
\resizebox{0.5\textwidth}{!}{
\begin{tabular}{lcccc}
\hline \noalign{\smallskip}
\rowcolor{Gray}
Model  & Art & Cartoons & Photos & Sketches  \\
\hline
ERM-ViT  & 0.908  & 0.882&	0.921& 	0.748\\
ERM-SDViT  & \textbf{0.948} &  \textbf{0.904}  & \textbf{0.950}  & \textbf{0.805} \\
\hline
\end{tabular}
}
\label{tab:cosine_table1}
\end{SCtable}

\noindent\textbf{SDViT with ERM based on ResNet:} Tab.~\ref{tab:resnet_50_SDViT_PACS} shows that our SD can also improve the performance of a competitive DG baseline based on ResNet-50.
\begin{SCtable}[][!htp]
\caption{\small Our self distillation (SD) improves the performance of ResNet-50 over DG baseline (ERM).}
\label{tab:resnet_50_SDViT_PACS}
\scalebox{0.73}{
\begin{tabular}{lccc}
\hline
\rowcolor{Gray}
 \multicolumn{2}{c}{ResNet50}&\multicolumn{2}{c}{DeiT-Small}\\
 \hline
 ERM & Ours (ERM-SD) & ERM & Ours (ERM-SD)  \\
\hline
 84.1 $\pm$ 0.4 &\textbf{ 85.7 $\pm$ 0.4 }& 84.9 $\pm$ 0.9 &\textbf{ 86.3 $\pm$ 0.2} \\
\hline
\end{tabular}}
\end{SCtable}

\noindent \textbf{On different block selection techniques:} We show performance with different ways of selecting blocks in our self-distillation method (see Table~ \ref{ablation_Distillation}). First, we restrict the random sampling of blocks to earlier blocks i.e. from block 0 to block 5. Second, we limit the random sampling of blocks to later blocks i.e. from block 6 to block 11. Finally, we do not perform any sampling in any range and rather include all the blocks \cite{zhang2019your}. Compared to all these block selection techniques, our proposal of random sampling from the full range of blocks shows the best (overall) average accuracy of 86.3\%. Sampling from earlier blocks seems beneficial as compared to the later blocks. When earlier blocks, with relatively longer feature pathways to the final block, are provided with soft target labels, there is potentially greater room for exploring cross-domain generalizable features.

\noindent \textbf{Confusion matrices:} Fig.~\ref{fig:confusion_matrix_PACS_average} visualizes the confusion matrices for the baseline and our method on \texttt{PACS} dataset. Compared to baseline, our method produces less number of false positives, particularly in `Sketches' as the target domain.

\noindent \textbf{What kind of features our DG approach facilitates?:} We visualize the features used by the baseline and our method to make predictions through visualizing attention maps (see Fig.~\ref{fig:attention_maps}). In all target domains, our method mostly attends to features that mainly capture the semantics and the shape of the object class. Whereas the baseline has a greater tendency to attend background features and thus focus less on the object class features. 

\noindent\textbf{Training overhead:} Table~\ref{comp_time} reports training overhead, computed as a relative \% increase in training time (hrs.), introduced by our method on top of the baseline. Our method adds very little training overhead over the baseline.

\noindent\textbf{Performance under different domain shifts:} We benchmark the performance under various domain shifts, including background shifts, corruption shifts, texture shifts, and style shifts. For instance, background shifts do not affect pixel, shape, texture, and structures in the foreground objects. Whereas style shifts typically depict variance at different stages of concepts, including texture, shape, and object part \cite{zhang2021delving}. To this end, we classify five DG datasets, including \texttt{PACS}, \texttt{VLCS}, \texttt{OfficeHome}, \texttt{TerraIncognita}, and \texttt{DomainNet}, into these four different domain shift categories based on the type of shift(s) exhibited by them. Table~\ref{taxonomy} reports the results by ERM-ViT (baseline) and ERM-SDViT (ours) under four different kinds of domain shifts. We observe that ERM-SDViT outperforms ERM-ViT across the whole spectrum of domain shifts. See Appendix for more results.  
\begin{SCtable}[][!t]
\caption{\small Training overhead, computed as relative \% increase in training time (hrs.), introduced by our method on top of the baseline.}
\resizebox{0.5\textwidth}{!}{
\begin{tabular}{lcccc}
\hline \noalign{\smallskip}
\rowcolor{Gray}
Model  & Art & Cartoons & Photos & Sketches  \\
\hline
ERM-ViT  &  0.266 & 0.270 &	0.278	& 0.267	  \\
ERM-SDViT  &  0.279 &  0.276  & 0.279 & 0.278 \\
\hline
Rel.overhead & 4.88 & 2.22 & 0.35 & 4.11 \\
\noalign{\smallskip}
\hline
\end{tabular}
}
\label{comp_time}
\end{SCtable}

\begin{table}[!htp]
\begin{center}
\caption{\small Performance under various domain shifts. Each entry is the accuracy (\%) averaged over the datasets belonging to a certain domain shift category.}

\label{taxonomy}
\adjustbox{max width=\textwidth}{
\begin{tabular}{lccccc}
\rowcolor{Gray}
\hline
\rowcolor{Gray}
\multirow{2}{*}{Methods} &  \multicolumn{5}{c}{ Shift Type } \\  \cline{2-6}   & &Background Shifts & ~~Corruption Shifts & Texture Shifts & Style Shifts \\  \rowcolor{Gray} \hline
& & \texttt{(VLCS,Terra)} & \texttt{(Terra)} & \texttt{(PACS,DomainNet)} & \texttt{(OH,PACS,DomainNet)} \\

ERM-ViT  & & 60.0  & 43.4 & 65.2	&67.2	  \\
ERM-SDViT & & \textbf{61.6} & \textbf{44.3}   &   \textbf{66.0}  &  \textbf{67.8}       \\

\noalign{\smallskip}
\hline
\end{tabular}
}
\end{center}
\end{table}
\section{Conclusion}

We propose a simple, plug-and-play approach namely self-distillation in ViTs for tackling DG. It provides soft supervision to the intermediate blocks of ViTs to strengthen their internal representations, thereby moderating the learning of input-output mapping problem. Extensive experiments on five datasets with different DG baselines and ViT backbones, including comparisons with the recent SOTA, validate the effectiveness of our approach for ViTs tackling DG problem.

\bibliographystyle{splncs}
\bibliography{egbib}

\begin{thebibliography}{10}

\bibitem{vaswani2017attention}
Vaswani, A., Shazeer, N., Parmar, N., Uszkoreit, J., Jones, L., Gomez, A.N.,
  Kaiser, {\L}., Polosukhin, I.:
\newblock Attention is all you need.
\newblock Advances in neural information processing systems \textbf{30} (2017)

\bibitem{devlin2018bert}
Devlin, J., Chang, M.W., Lee, K., Toutanova, K.:
\newblock Bert: Pre-training of deep bidirectional transformers for language
  understanding.
\newblock arXiv preprint arXiv:1810.04805 (2018)

\bibitem{brown2020language}
Brown, T., Mann, B., Ryder, N., Subbiah, M., Kaplan, J.D., Dhariwal, P.,
  Neelakantan, A., Shyam, P., Sastry, G., Askell, A.,  et~al.:
\newblock Language models are few-shot learners.
\newblock Advances in neural information processing systems \textbf{33} (2020)
  1877--1901

\bibitem{dosovitskiy2020image}
Dosovitskiy, A., Beyer, L., Kolesnikov, A., Weissenborn, D., Zhai, X.,
  Unterthiner, T., Dehghani, M., Minderer, M., Heigold, G., Gelly, S.,  et~al.:
\newblock An image is worth 16x16 words: Transformers for image recognition at
  scale.
\newblock arXiv preprint arXiv:2010.11929 (2020)

\bibitem{carion2020end}
Carion, N., Massa, F., Synnaeve, G., Usunier, N., Kirillov, A., Zagoruyko, S.:
\newblock End-to-end object detection with transformers.
\newblock In: European conference on computer vision, Springer (2020)  213--229

\bibitem{zhu2020deformable}
Zhu, X., Su, W., Lu, L., Li, B., Wang, X., Dai, J.:
\newblock Deformable detr: Deformable transformers for end-to-end object
  detection.
\newblock arXiv preprint arXiv:2010.04159 (2020)

\bibitem{zheng2021rethinking}
Zheng, S., Lu, J., Zhao, H., Zhu, X., Luo, Z., Wang, Y., Fu, Y., Feng, J.,
  Xiang, T., Torr, P.H.,  et~al.:
\newblock Rethinking semantic segmentation from a sequence-to-sequence
  perspective with transformers.
\newblock In: Proceedings of the IEEE/CVF conference on computer vision and
  pattern recognition. (2021)  6881--6890

\bibitem{wang2021pyramid}
Wang, W., Xie, E., Li, X., Fan, D.P., Song, K., Liang, D., Lu, T., Luo, P.,
  Shao, L.:
\newblock Pyramid vision transformer: A versatile backbone for dense prediction
  without convolutions.
\newblock In: Proceedings of the IEEE/CVF International Conference on Computer
  Vision. (2021)  568--578

\bibitem{muandet2013domain}
Muandet, K., Balduzzi, D., Sch{\"o}lkopf, B.:
\newblock Domain generalization via invariant feature representation.
\newblock In: International Conference on Machine Learning. (2013)  10--18

\bibitem{ghifary2015domain}
Ghifary, M., Bastiaan~Kleijn, W., Zhang, M., Balduzzi, D.:
\newblock Domain generalization for object recognition with multi-task
  autoencoders.
\newblock In: Proceedings of the IEEE international conference on computer
  vision. (2015)  2551--2559

\bibitem{li2017deeper}
Li, D., Yang, Y., Song, Y.Z., Hospedales, T.M.:
\newblock Deeper, broader and artier domain generalization.
\newblock In: Proceedings of the IEEE international conference on computer
  vision. (2017)  5542--5550

\bibitem{carlucci2019domain}
Carlucci, F.M., D'Innocente, A., Bucci, S., Caputo, B., Tommasi, T.:
\newblock Domain generalization by solving jigsaw puzzles.
\newblock In: Proceedings of the IEEE Conference on Computer Vision and Pattern
  Recognition. (2019)  2229--2238

\bibitem{Gulrajani2021InSO}
Gulrajani, I., Lopez-Paz, D.:
\newblock In search of lost domain generalization.
\newblock ArXiv \textbf{abs/2007.01434} (2021)

\bibitem{huang2020self}
Huang, Z., Wang, H., Xing, E.P., Huang, D.:
\newblock Self-challenging improves cross-domain generalization.
\newblock (2020)

\bibitem{khan2021mode}
Khan, M.H., Zaidi, T., Khan, S., Khan, F.S.:
\newblock Mode-guided feature augmentation for domain generalization.
\newblock (2021)

\bibitem{Nam_2021_CVPR}
Nam, H., Lee, H., Park, J., Yoon, W., Yoo, D.:
\newblock Reducing domain gap by reducing style bias.
\newblock In: Proceedings of the IEEE/CVF Conference on Computer Vision and
  Pattern Recognition (CVPR). (2021)  8690--8699

\bibitem{Kim_2021_ICCV}
Kim, D., Yoo, Y., Park, S., Kim, J., Lee, J.:
\newblock Selfreg: Self-supervised contrastive regularization for domain
  generalization.
\newblock In: Proceedings of the IEEE/CVF International Conference on Computer
  Vision (ICCV). (2021)  9619--9628

\bibitem{bui2021exploiting}
Bui, M.H., Tran, T., Tran, A., Phung, D.:
\newblock Exploiting domain-specific features to enhance domain generalization.
\newblock Advances in Neural Information Processing Systems \textbf{34} (2021)

\bibitem{ganin2016domain}
Ganin, Y., Ustinova, E., Ajakan, H., Germain, P., Larochelle, H., Laviolette,
  F., Marchand, M., Lempitsky, V.:
\newblock Domain-adversarial training of neural networks.
\newblock The Journal of Machine Learning Research \textbf{17} (2016)
  2096--2030

\bibitem{li2018domain}
Li, H., Pan, S.J., Wang, S., Kot, A.C.:
\newblock Domain generalization with adversarial feature learning.
\newblock In: Proceedings of the IEEE conference on computer vision and pattern
  recognition. (2018)  5400--5409

\bibitem{li2018learning}
Li, D., Yang, Y., Song, Y.Z., Hospedales, T.M.:
\newblock Learning to generalize: Meta-learning for domain generalization.
\newblock In: Thirty-Second AAAI Conference on Artificial Intelligence. (2018)

\bibitem{dou2019domain}
Dou, Q., de~Castro, D.C., Kamnitsas, K., Glocker, B.:
\newblock Domain generalization via model-agnostic learning of semantic
  features.
\newblock In: Advances in Neural Information Processing Systems. (2019)
  6450--6461

\bibitem{balaji2018metareg}
Balaji, Y., Sankaranarayanan, S., Chellappa, R.:
\newblock Metareg: Towards domain generalization using meta-regularization.
\newblock In: Advances in Neural Information Processing Systems. (2018)
  998--1008

\bibitem{li2019episodic}
Li, D., Zhang, J., Yang, Y., Liu, C., Song, Y.Z., Hospedales, T.M.:
\newblock Episodic training for domain generalization.
\newblock In ICCV (2019)

\bibitem{shankar2018generalizing}
Shankar, S., Piratla, V., Chakrabarti, S., Chaudhuri, S., Jyothi, P., Sarawagi,
  S.:
\newblock Generalizing across domains via cross-gradient training.
\newblock arXiv preprint arXiv:1804.10745 (2018)

\bibitem{volpi2018generalizing}
Volpi, R., Namkoong, H., Sener, O., Duchi, J.C., Murino, V., Savarese, S.:
\newblock Generalizing to unseen domains via adversarial data augmentation.
\newblock In: Advances in Neural Information Processing Systems. (2018)
  5334--5344

\bibitem{zhou2020learning}
Zhou, K., Yang, Y., Hospedales, T., Xiang, T.:
\newblock Learning to generate novel domains for domain generalization.
\newblock (2020)

\bibitem{wang2020learning}
Wang, S., Yu, L., Li, C., Fu, C.W., Heng, P.A.:
\newblock Learning from extrinsic and intrinsic supervisions for domain
  generalization.
\newblock (2020)

\bibitem{naseer2021improving}
Naseer, M., Ranasinghe, K., Khan, S., Khan, F.S., Porikli, F.:
\newblock On improving adversarial transferability of vision transformers.
\newblock arXiv preprint arXiv:2106.04169 (2021)

\bibitem{cha2021swad}
Cha, J., Chun, S., Lee, K., Cho, H.C., Park, S., Lee, Y., Park, S.:
\newblock Swad: Domain generalization by seeking flat minima.
\newblock Advances in Neural Information Processing Systems \textbf{34} (2021)

\bibitem{vapnik1999nature}
Vapnik, V.:
\newblock The nature of statistical learning theory.
\newblock Springer science \& business media (1999)

\bibitem{yang2013multi}
YANG, P.Y., Gao, W.:
\newblock Multi-view discriminant transfer learning.
\newblock (2013)

\bibitem{arjovsky2019invariant}
Arjovsky, M., Bottou, L., Gulrajani, I., Lopez-Paz, D.:
\newblock Invariant risk minimization.
\newblock arXiv preprint arXiv:1907.02893 (2019)

\bibitem{xu2014exploiting}
Xu, Z., Li, W., Niu, L., Xu, D.:
\newblock Exploiting low-rank structure from latent domains for domain
  generalization.
\newblock In: European Conference on Computer Vision, Springer (2014)  628--643

\bibitem{chattopadhyay2020learning}
Chattopadhyay, P., Balaji, Y., Hoffman, J.:
\newblock Learning to balance specificity and invariance for in and out of
  domain generalization.
\newblock (2020)

\bibitem{seo2019learning}
Seo, S., Suh, Y., Kim, D., Kim, G., Han, J., Han, B.:
\newblock Learning to optimize domain specific normalization for domain
  generalization.
\newblock (2020)

\bibitem{motiian2017unified}
Motiian, S., Piccirilli, M., Adjeroh, D.A., Doretto, G.:
\newblock Unified deep supervised domain adaptation and generalization.
\newblock In: Proceedings of the IEEE International Conference on Computer
  Vision. (2017)  5715--5725

\bibitem{wu2021cvt}
Wu, H., Xiao, B., Codella, N., Liu, M., Dai, X., Yuan, L., Zhang, L.:
\newblock Cvt: Introducing convolutions to vision transformers.
\newblock In: Proceedings of the IEEE/CVF International Conference on Computer
  Vision. (2021)  22--31

\bibitem{touvron2021training}
Touvron, H., Cord, M., Douze, M., Massa, F., Sablayrolles, A., J{\'e}gou, H.:
\newblock Training data-efficient image transformers \& distillation through
  attention.
\newblock In: International Conference on Machine Learning, PMLR (2021)
  10347--10357

\bibitem{Dai_2021_ICCV}
Dai, X., Chen, Y., Yang, J., Zhang, P., Yuan, L., Zhang, L.:
\newblock Dynamic detr: End-to-end object detection with dynamic attention.
\newblock In: Proceedings of the IEEE/CVF International Conference on Computer
  Vision (ICCV). (2021)  2988--2997

\bibitem{strudel2021segmenter}
Strudel, R., Garcia, R., Laptev, I., Schmid, C.:
\newblock Segmenter: Transformer for semantic segmentation.
\newblock In: Proceedings of the IEEE/CVF International Conference on Computer
  Vision. (2021)  7262--7272

\bibitem{lu2021simpler}
Lu, Z., He, S., Zhu, X., Zhang, L., Song, Y.Z., Xiang, T.:
\newblock Simpler is better: Few-shot semantic segmentation with classifier
  weight transformer.
\newblock In: Proceedings of the IEEE/CVF International Conference on Computer
  Vision. (2021)  8741--8750

\bibitem{yuan2021tokens}
Yuan, L., Chen, Y., Wang, T., Yu, W., Shi, Y., Jiang, Z.H., Tay, F.E., Feng,
  J., Yan, S.:
\newblock Tokens-to-token vit: Training vision transformers from scratch on
  imagenet.
\newblock In: Proceedings of the IEEE/CVF International Conference on Computer
  Vision. (2021)  558--567

\bibitem{zhang2021delving}
Zhang, C., Zhang, M., Zhang, S., Jin, D., Zhou, Q., Cai, Z., Zhao, H., Yi, S.,
  Liu, X., Liu, Z.:
\newblock Delving deep into the generalization of vision transformers under
  distribution shifts.
\newblock arXiv preprint arXiv:2106.07617 (2021)

\bibitem{hinton2015distilling}
Hinton, G., Vinyals, O., Dean, J.:
\newblock Distilling the knowledge in a neural network (2015).
\newblock arXiv preprint arXiv:1503.02531 \textbf{2} (2015)

\bibitem{zhang2019your}
Zhang, L., Song, J., Gao, A., Chen, J., Bao, C., Ma, K.:
\newblock Be your own teacher: Improve the performance of convolutional neural
  networks via self distillation.
\newblock In: Proceedings of the IEEE/CVF International Conference on Computer
  Vision. (2019)  3713--3722

\bibitem{yun2020regularizing}
Yun, S., Park, J., Lee, K., Shin, J.:
\newblock Regularizing class-wise predictions via self-knowledge distillation.
\newblock In: Proceedings of the IEEE/CVF conference on computer vision and
  pattern recognition. (2020)  13876--13885

\bibitem{wang2021embracing}
Wang, Y., Li, H., Chau, L.p., Kot, A.C.:
\newblock Embracing the dark knowledge: Domain generalization using regularized
  knowledge distillation.
\newblock In: Proceedings of the 29th ACM International Conference on
  Multimedia. (2021)  2595--2604

\bibitem{fang2013unbiased}
Fang, C., Xu, Y., Rockmore, D.N.:
\newblock Unbiased metric learning: On the utilization of multiple datasets and
  web images for softening bias.
\newblock In: Proceedings of the IEEE International Conference on Computer
  Vision. (2013)  1657--1664

\bibitem{venkateswara2017deep}
Venkateswara, H., Eusebio, J., Chakraborty, S., Panchanathan, S.:
\newblock Deep hashing network for unsupervised domain adaptation.
\newblock In: Proceedings of the IEEE Conference on Computer Vision and Pattern
  Recognition. (2017)  5018--5027

\bibitem{beery2018recognition}
Beery, S., Van~Horn, G., Perona, P.:
\newblock Recognition in terra incognita.
\newblock In: Proceedings of the European conference on computer vision (ECCV).
  (2018)  456--473

\bibitem{peng2019moment}
Peng, X., Bai, Q., Xia, X., Huang, Z., Saenko, K., Wang, B.:
\newblock Moment matching for multi-source domain adaptation.
\newblock In: Proceedings of the IEEE International Conference on Computer
  Vision. (2019)  1406--1415

\bibitem{loshchilov2018fixing}
Loshchilov, I., Hutter, F.:
\newblock Fixing weight decay regularization in adam.
\newblock (2018)

\bibitem{sagawa2019distributionally}
Sagawa, S., Koh, P.W., Hashimoto, T.B., Liang, P.:
\newblock Distributionally robust neural networks for group shifts: On the
  importance of regularization for worst-case generalization.
\newblock arXiv preprint arXiv:1911.08731 (2019)

\bibitem{yan2020improve}
Yan, S., Song, H., Li, N., Zou, L., Ren, L.:
\newblock Improve unsupervised domain adaptation with mixup training.
\newblock arXiv preprint arXiv:2001.00677 (2020)

\bibitem{sun2016deep}
Sun, B., Saenko, K.:
\newblock Deep coral: Correlation alignment for deep domain adaptation.
\newblock In: European conference on computer vision, Springer (2016)  443--450

\bibitem{li2018deep}
Li, Y., Tian, X., Gong, M., Liu, Y., Liu, T., Zhang, K., Tao, D.:
\newblock Deep domain generalization via conditional invariant adversarial
  networks.
\newblock In: Proceedings of the European Conference on Computer Vision (ECCV).
  (2018)  624--639

\bibitem{blanchard2017domain}
Blanchard, G., Deshmukh, A.A., Dogan, U., Lee, G., Scott, C.:
\newblock Domain generalization by marginal transfer learning.
\newblock arXiv preprint arXiv:1711.07910 (2017)

\bibitem{zhang2021adaptive}
Zhang, M., Marklund, H., Dhawan, N., Gupta, A., Levine, S., Finn, C.:
\newblock Adaptive risk minimization: Learning to adapt to domain shift.
\newblock Advances in Neural Information Processing Systems \textbf{34} (2021)

\bibitem{krueger2021out}
Krueger, D., Caballero, E., Jacobsen, J.H., Zhang, A., Binas, J., Zhang, D.,
  Le~Priol, R., Courville, A.:
\newblock Out-of-distribution generalization via risk extrapolation (rex).
\newblock In: International Conference on Machine Learning, PMLR (2021)
  5815--5826

\bibitem{iwasawa2021test}
Iwasawa, Y., Matsuo, Y.:
\newblock Test-time classifier adjustment module for model-agnostic domain
  generalization.
\newblock Advances in Neural Information Processing Systems \textbf{34} (2021)

\end{thebibliography}
\clearpage
\section*{Supplementary Material} \label{t-SNE_appendex}
\begin{figure}
  \centering
  \includegraphics[width=\textwidth]{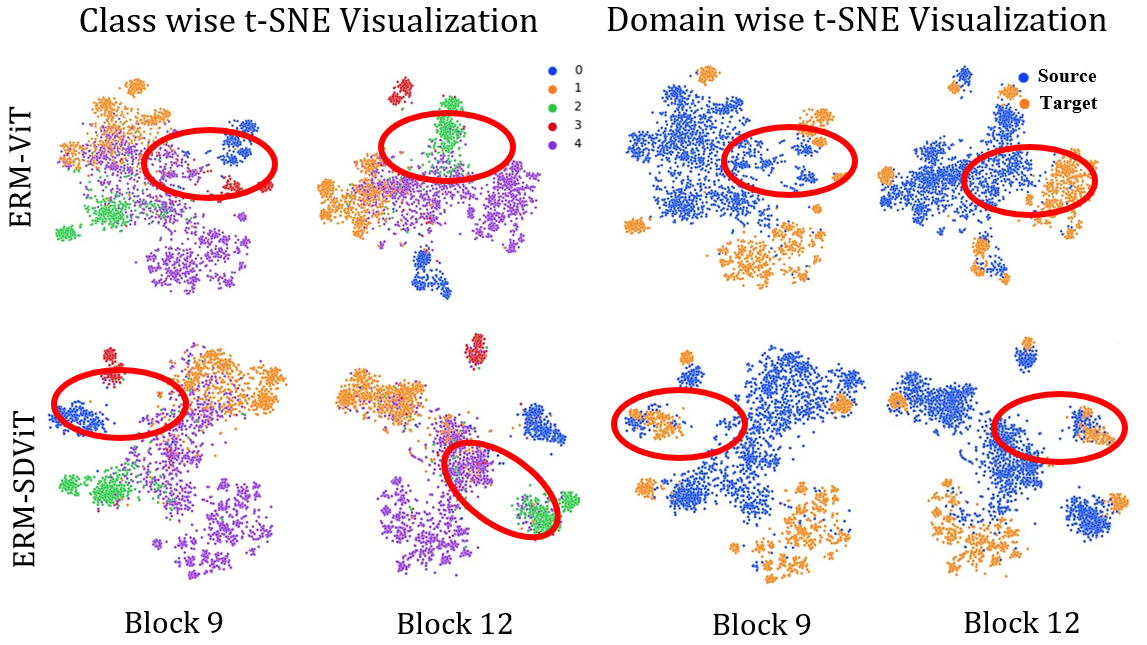}
  \caption{\small t-SNE visualization of features from different blocks (9 \& 12) in baseline (ERM-ViT) and our approach for Caltech101 target domain in \texttt{VLCS} dataset. Left: Features are colored corresponding to their class labels (classes: 5). Right: Features are colored corresponding to their domain labels. Our approach (ERM-SDViT) improves class-wise discrimination. For instance, in class-wise t-SNE in block 9, the features of class 0 and 3 (highlighted in red circle) are well-separated as compared to the baseline (ERM-ViT). Similarly, in block 12, the features of class 2 and 4 are clearly distinguishable. In domain-wise t-SNE, for our approach, source and target domain features show greater overlap with each other.}
  \label{fig:Tsne_cls_VLCS}
\end{figure}

\noindent \textbf{More t-SNE feature visualizations:} Fig.~\ref{fig:Tsne_cls_VLCS} (left) visualizes the class-wise feature representations of different blocks using t-SNE in baseline (ERM-ViT) and our model (ERM-SDViT) for Caltech101 target domain in the \texttt{VLCS} dataset. In comparison to baseline, our method facilitates improved learning of discriminative features and hence reduces the intra-class variance while increasing the inter-class variance in the feature space. 
Similarly, Fig. \ref{fig:Tsne_cls_VLCS} (right) visualizes the same features, however, on the basis of source and target domain labels. Compared to baseline, our method promotes a greater overlap between the features of source and target domain features. 

\noindent \textbf{Hyperparameters analysis:} We show test performance as a function of temperature ($\tau$) and weight $\lambda$ (Tab.~ \ref{HP_analysis_Table}). Note that, we consider the final output (in our all experimental results) which is obtained as the highest validation accuracy via grid search. The individual results of hyper-parameters mentioned in the Tab.~ \ref{HP_analysis_Table} could not be considered best as they show output on the test data, hence violating the DG protocols of model performance on unseen target data. 
\begin{table}[!htp]
\begin{center}
\caption{\small Analysis of temperature ($\tau$) and weight $\lambda$ with CvT-21 on PACS.}
\setlength{\tabcolsep}{3pt}
\adjustbox{max width=\textwidth}{
\begin{tabular}{lcccccc}
\hline
\rowcolor{Gray}
$\tau$, $\lambda$=3.0, 0.1 &$\tau$, $\lambda$=3.0, 0.2  & $\tau$, $\lambda$=3.0, 0.5 & $\tau$, $\lambda$=5.0, 0.1 & $\tau$, $\lambda$=5.0, 0.2& $\tau$, $\lambda$=5.0, 0.5 \\
\hline
88.2 $\pm$ 0.3 & 88.2 $\pm$ 0.4 & 87.2 $\pm$ 0.5 & 88.4 $\pm$ 0.1 & \textbf{89.7 $\pm$ 0.7} & 88.5 $\pm$ 0.4 & \\
\hline
\end{tabular}}
\label{HP_analysis_Table} 
\end{center}
\end{table}

\begin{figure}
  \centering
  \includegraphics[width=\textwidth]{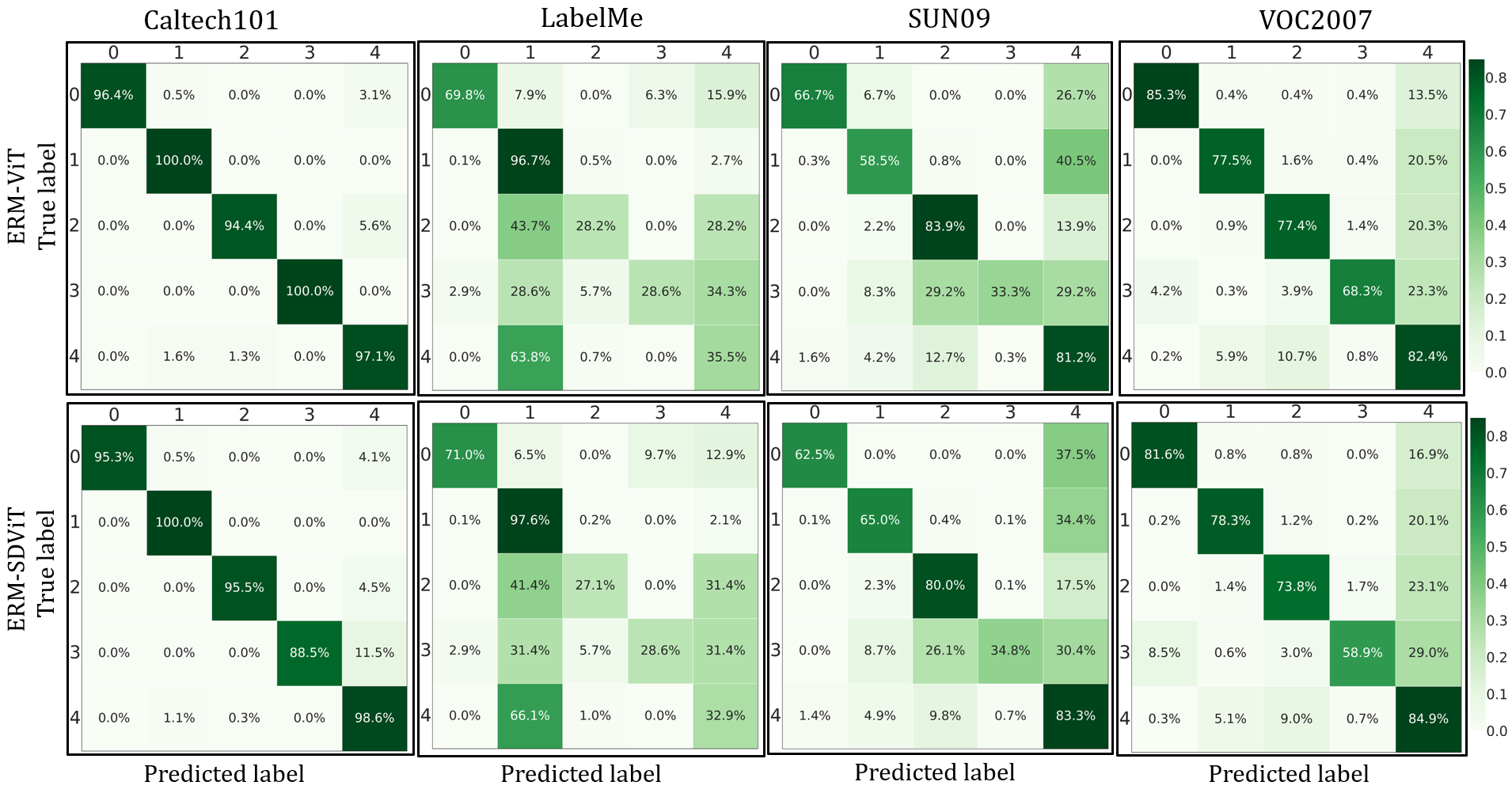}
  \caption{\small Confusion matrices of the baseline and our method on \texttt{VLCS} dataset. The classes in the figure are `0':Bird, `1':Car, `2':Chair, `3':Dog, and `4':Person.} 
  \label{fig:confusion_matrix_VLCS}
\end{figure}

\noindent \textbf{Confusion matrices on other DG dataset:} Fig.~\ref{fig:confusion_matrix_VLCS} visualizes the confusion matrices for the baseline and our method on \texttt{VLCS} dataset. In comparison to the baseline, our method is capable of reducing false positives in all four target domains. 

\begin{SCfigure}[][!t]
\caption{\small Comparison of attention maps between the baseline (ERM-ViT) and our proposed method (ERM-SDViT) on four target domains of \texttt{TerraIncognita} dataset. The ViT backbone is DeiT-Small.}
\resizebox{0.6\textwidth}{!}{
  \includegraphics[width=\textwidth]{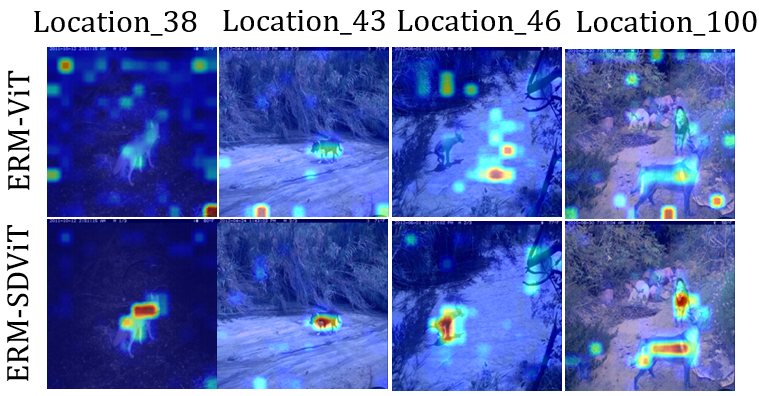}}
  \label{fig:Attentions_Terra}
\end{SCfigure}

\begin{figure}[!t]
  \centering
  \includegraphics[width=\textwidth]{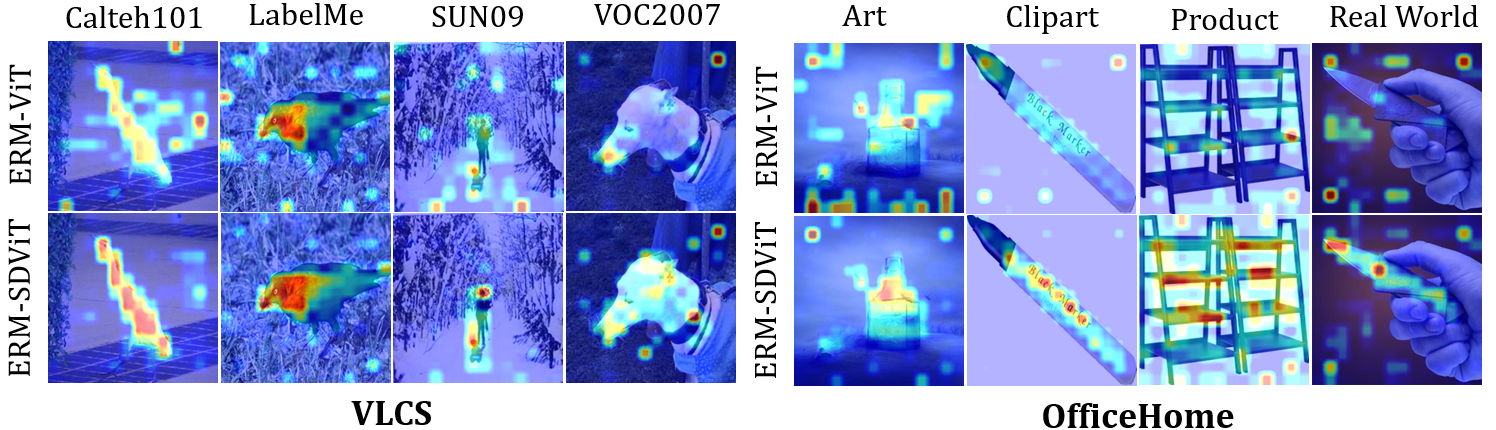}
  \caption{\small Comparison of attention maps between the baseline (ERM-ViT) and our proposed method (ERM-SDViT) on four target domains of \texttt{VLCS} and \texttt{OfficeHome} datasets. The ViT backbone is DeiT-Small.} 
\label{fig:attention_maps_VLCS_OH} 
\end{figure}

\begin{figure}[!t]
  \centering
  \includegraphics[width=\textwidth]{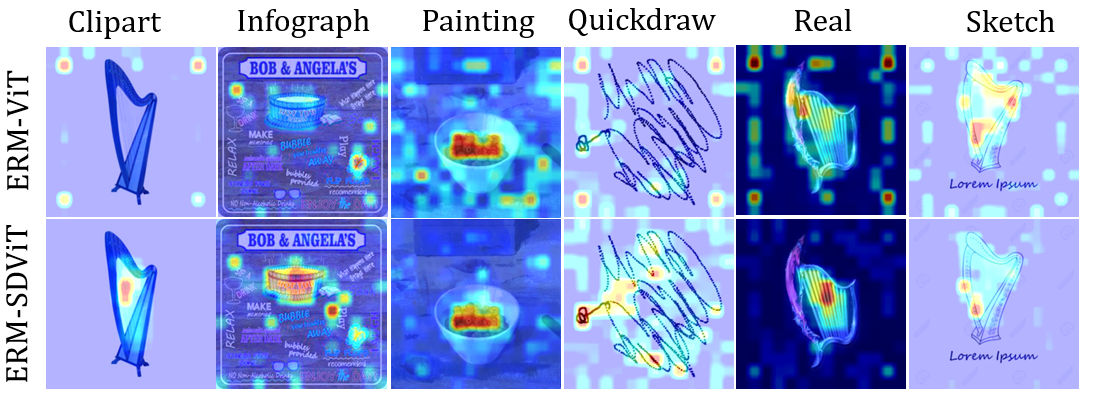}
  \caption{\small Comparison of attention maps between the baseline (ERM-ViT) and our proposed method (ERM-SDViT) on six target domains of \texttt{DomainNet} datasets. The ViT backbone is DeiT-Small.} 
\label{fig:attention_maps_DomainNet}
\end{figure}

\noindent \textbf{Attention visualizations on other DG datasets:} We also visualize attention maps from different images of four datasets, including \texttt{VLCS}, \texttt{OfficeHome}, \texttt{TerraIncognita} and \texttt{DomainNet} in Figs.~\ref{fig:Attentions_Terra}, \ref{fig:attention_maps_VLCS_OH} and \ref{fig:attention_maps_DomainNet}. It can be observed 
that in all target domains of the four datasets, our method mostly relies on features corresponding to the foreground object's semantics rather than the background information. However, the baseline approach (ERM-ViT) mostly capitalizes more on the background features and pays less attention to the features belonging to the foreground object. 
For instance, in Fig.~\ref{fig:Attentions_Terra}, target domain: Location\_46 of the \texttt{TerraIncognita} dataset, our method is capable of focussing on the foreground object (a dog), which occupies a small fraction of the overall image. However, the baseline model is prone to attending more to the background features, which are prevalent in the image.
Note that the attention maps are computed at the final block of ViT models.


\noindent \textbf{Recognition accuracy on target domains of other DG datasets:} Tables~\ref{tab:ablation_VLCS_OH_Terra} and \ref{tab:ablation_DomainNet} compares target domain-wise recognition accuracy on \texttt{VLCS}, \texttt{OfficeHome}, \texttt{TerraIncognita}, and \texttt{DomainNet} datasets of our method with the baseline utilizing three ViT backbones and a DG baseline (T3A~\cite{iwasawa2021test}).

\begin{table}
\begin{center}
\caption{\small Comparison of target domain-wise classification accuracy on \texttt{VLCS}, \texttt{OfficeHome}, and \texttt{TerraIncognita} datasets. Results are reported of our method with the baseline using three different ViT backbones, including DeiT-Small \cite{touvron2021training}, CvT-21 \cite{wu2021cvt}, and T2T-ViT-14 \cite{yuan2021tokens}, and a DG baseline (T3A~\cite{iwasawa2021test}).}

\label{tab:ablation_VLCS_OH_Terra}
\adjustbox{max width=\textwidth}{
\begin{tabular}{lccccccc}
\hline \noalign{\smallskip}
\rowcolor{Gray}
Dataset &   & & \texttt{VLCS} & &  & &\\
\noalign{\smallskip}
\hline
\rowcolor{Gray} Model & Backbone & \# of Params & Caltech101 & LableMe & SUN09 & VOC2007 & Average \\
\noalign{\smallskip}
\hline
\noalign{\smallskip}
ERM  & ResNet-50 & 23.5M & 98.1 $\pm$ 0.4   &       64.1 $\pm$ 0.5     &     70.7 $\pm$ 0.9       &   74.8 $\pm$ 2.4 & 76.9 $\pm$ 0.6   \\
ERM-ViT  & DeiT-Small & 22M & 96.7 $\pm$ 0.8     &     65.2 $\pm$ 1.0      &    73.9 $\pm$ 0.3     &     77.4 $\pm$ 0.3 & 78.3 $\pm$ 0.5  \\
ERM-SDViT & DeiT-Small & 22M & 96.8 $\pm$ 0.5      &    64.2 $\pm$ 0.8   &       76.2 $\pm$ 0.4      &    78.5 $\pm$ 0.4 & 78.9 $\pm$ 0.4     \\
ERM-SDViT + T3A & DeiT-Small & 22M & 98.9 $\pm$ 0.2  &        65.9 $\pm$ 0.3     &     79.8 $\pm$ 0.4    &      81.9 $\pm$ 0.4 & \underline{81.6 $\pm$ 0.1 }\\
ERM-ViT  & CvT-21 & 32M &   97.3 $\pm$ 0.5 &   65.2 $\pm$ 0.9  &    76.6 $\pm$ 1.1 &     76.9 $\pm$ 0.3 & 79.0 $\pm$ 0.3  \\
ERM-SDViT & CvT-21 & 32M &  96.5 $\pm$ 0.7  &        63.3 $\pm$ 0.4      &    78.1 $\pm$ 0.2     &     78.9 $\pm$ 0.8 & 79.2 $\pm$ 0.4       \\
ERM-SDViT + T3A & CvT-21 & 32M  &  98.4 $\pm$ 0.3    &      66.8 $\pm$ 0.5   &       80.1 $\pm$ 1.0    &      80.6 $\pm$ 0.7  & \textbf{81.9 $\pm$ 0.4}   \\
ERM-ViT  & T2T-ViT-14 &  21.5M & 96.5 $\pm$ 0.5 & 64.5 $\pm$ 0.1     &     76.4 $\pm$ 0.4 &         78.2 $\pm$ 1.0 & 78.9 $\pm$ 0.3   \\
ERM-SDViT & T2T-ViT-14 & 21.5M &  96.9 $\pm$ 0.4    &  64.0 $\pm$ 0.5  &        76.7 $\pm$ 1.4   &  80.4 $\pm$ 1.3    & 79.5 $\pm$ 0.8   \\
ERM-SDViT + T3A & T2T-ViT-14 & 21.5M & 98.6 $\pm$ 0.3   &       66.5 $\pm$ 0.7     &     78.2 $\pm$ 0.5   &       81.7 $\pm$ 0.9      &  81.2 $\pm$ 0.3   \\
\hline
\rowcolor{Gray}
Dataset &   & & \texttt{OfficeHome} & &  & &\\
\noalign{\smallskip}
\hline
\rowcolor{Gray} Model & Backbone & \# of Params & Art & Clipart & Product & Real World & Average \\
\noalign{\smallskip}
\hline
\noalign{\smallskip}
ERM  & ResNet-50 & 23.5M & 58.8 $\pm$ 1.0  &        51.3 $\pm$ 0.4    &      73.7 $\pm$ 0.4    &      74.7 $\pm$ 0.6 & 64.6 $\pm$ 0.2   \\
ERM-ViT  & DeiT-Small & 22M &  67.6 $\pm$ 0.3    &57.0 $\pm$ 0.6 &     79.4 $\pm$ 0.1    & 81.6 $\pm$ 0.4 & 71.4 $\pm$ 0.1    \\
ERM-SDViT & DeiT-Small & 22M & 68.3 $\pm$ 0.8   &  56.3 $\pm$ 0.2   & 79.5 $\pm$ 0.3  &    81.8 $\pm$ 0.1   &  71.5 $\pm$ 0.2    \\
ERM-SDViT + T3A & DeiT-Small & 22M & 69.1 $\pm$ 1.0  & 57.9 $\pm$ 0.4     &     80.7 $\pm$ 0.0      &    82.3 $\pm$ 0.1    & 72.5 $\pm$ 0.3  \\
ERM-ViT  & CvT-21 & 32M &   74.4 $\pm$ 0.2    &  59.8 $\pm$ 0.5      &    83.5 $\pm$ 0.4     &     84.1 $\pm$ 0.2 &  75.5 $\pm$ 0.0 \\
ERM-SDViT & CvT-21 & 32M & 73.8 $\pm$ 0.6  &    60.7 $\pm$ 0.9  & 83.0 $\pm$ 0.3   &       85.0 $\pm$ 0.3   & \underline{75.6 $\pm$ 0.2}    \\
ERM-SDViT + T3A & CvT-21 & 32M  & 75.2 $\pm$ 0.7     &     62.7 $\pm$ 0.8    &      84.2 $\pm$ 0.6  &        86.1 $\pm$ 0.0  &  \textbf{77.0 $\pm$ 0.2}  \\
ERM-ViT  & T2T-ViT-14 &  21.5M &  70.2 $\pm$ 0.5  &        59.0 $\pm$ 0.6      &    81.9 $\pm$ 0.3   &       83.6 $\pm$ 0.6 &  73.7 $\pm$ 0.2  \\
ERM-SDViT & T2T-ViT-14 & 21.5M &  71.1 $\pm$ 0.5      &    59.2 $\pm$ 0.3      &    82.8 $\pm$ 0.4      &    83.5 $\pm$ 0.3      & 74.2 $\pm$ 0.3  \\
ERM-SDViT + T3A & T2T-ViT-14 & 21.5M & 70.8 $\pm$ 0.4   &  61.9 $\pm$ 0.7  & 84.1 $\pm$ 0.2    &   85.0 $\pm$ 0.3 &  75.5 $\pm$ 0.2    \\
\hline
\rowcolor{Gray}
Dataset &   & & \texttt{TerraIncognita} & &  & &\\
\noalign{\smallskip}
\hline
\rowcolor{Gray} Model & Backbone & \# of Params & location\_38 & location\_43 & location\_46 & location\_100 & Average \\
\noalign{\smallskip}
\hline
\noalign{\smallskip}
ERM  & ResNet-50 & 23.5M & 56.3 $\pm$ 1.1    &  36.8 $\pm$ 4.6   &   52.6 $\pm$ 0.4   &   35.2 $\pm$ 1.7 & 45.2 $\pm$ 1.2  \\
ERM-ViT  & DeiT-Small & 22M & 50.2 $\pm$ 1.4 & 30.6 $\pm$ 0.9& 53.2 $\pm$ 0.2 & 39.6 $\pm$ 1.0 &43.4 $\pm$ 0.5  \\
ERM-SDViT & DeiT-Small & 22M & 55.9 $\pm$ 1.7   &       31.7 $\pm$ 2.6    &      52.2 $\pm$ 0.3  &        37.4 $\pm$ 0.6  &   44.3 $\pm$ 1.0   \\
ERM-SDViT + T3A & DeiT-Small & 22M & 53.8 $\pm$ 1.2 &    36.2 $\pm$ 1.0 &    51.1 $\pm$ 1.0 &       38.5 $\pm$ 1.3      &  44.9 $\pm$ 0.4  \\
ERM-ViT  & CvT-21 & 32M &   51.4 $\pm$ 1.8 & 40.1 $\pm$ 1.7&  57.6 $\pm$ 1.0 &45.7 $\pm$ 0.6 & 48.7 $\pm$ 0.4 \\
ERM-SDViT & CvT-21 & 32M & 53.6 $\pm$ 3.3   &       42.7 $\pm$ 1.6     &     58.2 $\pm$ 1.0   &       44.5 $\pm$ 1.8 & 49.7 $\pm$ 1.4   \\
ERM-SDViT + T3A & CvT-21 & 32M  &    58.1 $\pm$ 0.7     &     46.2 $\pm$ 0.3&  57.0 $\pm$ 1.0      &    44.1 $\pm$ 2.2 & \textbf{51.4 $\pm$ 0.7} \\
ERM-ViT  & T2T-ViT-14 &  21.5M &52.5 $\pm$ 1.7      &    43.0 $\pm$ 1.3   &       53.7 $\pm$ 1.1  &        43.0 $\pm$ 1.6         &  48.1 $\pm$ 0.2    \\
ERM-SDViT & T2T-ViT-14 & 21.5M & 57.2 $\pm$ 2.9       &   45.4 $\pm$ 2.4     &     57.7 $\pm$ 0.8  &        41.9 $\pm$ 0.4    & \underline{50.6 $\pm$ 0.8}   \\
ERM-SDViT + T3A & T2T-ViT-14 & 21.5M &59.3 $\pm$ 1.2    &      48.2 $\pm$ 1.0 &53.1 $\pm$ 0.9     &     41.5 $\pm$ 0.2      &  50.5 $\pm$ 0.6   \\
\hline
\end{tabular}
}
\end{center}
\end{table}

\begin{table}
\begin{center}
\caption{\small Comparison of target domain-wise classification accuracy on \texttt{DomainNet} dataset. Results are reported of our method with the baseline using three different ViT backbones, including DeiT-Small \cite{touvron2021training}, CvT-21 \cite{wu2021cvt}, and T2T-ViT-14 \cite{yuan2021tokens}, and a DG baseline (T3A~\cite{iwasawa2021test}).}
\label{tab:ablation_DomainNet}
\adjustbox{max width=\textwidth}{
\begin{tabular}{lcccccccccc}
\noalign{\smallskip}
\hline 
\rowcolor{Gray} Dataset &   & & &\texttt{DomainNet}  & & & & & & \\
\noalign{\smallskip}
\hline
\rowcolor{Gray} Model & Backbone & \# of Params & Clipart & Infograph  & Painting &  Quickdraw & Real & Sketch & Average \\
\noalign{\smallskip}
\hline
\noalign{\smallskip}
ERM  & ResNet-50 & 23.5M & 57.6 $\pm$ 0.6   &  18.5 $\pm$ 0.3 &  45.9 $\pm$ 0.7 & 11.6 $\pm$ 0.1 & 59.5 $\pm$ 0.3    &      48.6 $\pm$ 0.3 & 40.3 $\pm$ 0.1  \\
ERM-ViT  & DeiT-Small & 22M &62.9 $\pm$ 0.2   &       23.3 $\pm$ 0.1    &      53.1 $\pm$ 0.2   &       15.7 $\pm$ 0.1    &      65.7 $\pm$ 0.1    &      52.4 $\pm$ 0.2 & 45.5 $\pm$ 0.0  \\
ERM-SDViT & DeiT-Small & 22M & 63.4 $\pm$ 0.1   &       22.9 $\pm$ 0.0     &     53.7 $\pm$ 0.1    &      15.0 $\pm$ 0.4      &    67.4 $\pm$ 0.1       &   52.6 $\pm$ 0.2  &  45.8 $\pm$ 0.0 \\
ERM-SDViT + T3A & DeiT-Small & 22M &  64.3 $\pm$ 0.2    &      23.7 $\pm$ 0.0  & 54.2 $\pm$ 0.3    &      19.7 $\pm$ 0.4  & 69.6 $\pm$ 0.1    &      53.2 $\pm$ 0.2     & 47.4 $\pm$ 0.1  \\
ERM-ViT  & CvT-21 & 32M & 69.0 $\pm$ 0.2        &  27.2 $\pm$ 0.2    &      58.4 $\pm$ 0.2    &      17.1 $\pm$ 0.3    &      71.6 $\pm$ 0.1    &      59.2 $\pm$ 0.3   & 50.4 $\pm$ 0.1  \\
ERM-SDViT & CvT-21 & 32M & 68.9 $\pm$ 0.1      &    26.7 $\pm$ 0.3   &       58.0 $\pm$ 0.1   &       17.3 $\pm$ 0.1    &      71.9 $\pm$ 0.0     &     59.1 $\pm$ 0.3    & \underline{50.4 $\pm$ 0.0}   \\
ERM-SDViT + T3A & CvT-21 & 32M  & 69.7 $\pm$ 0.1    &      27.6 $\pm$ 0.2  &        58.7 $\pm$ 0.1     &     23.0 $\pm$ 0.1  &        73.6 $\pm$ 0.2    &      59.6 $\pm$ 0.1   &   \textbf{52.0 $\pm$ 0.0 } \\
ERM-ViT  & T2T-ViT-14 &  21.5M & 67.0 $\pm$ 0.3       &   25.2 $\pm$ 0.2   &       55.3 $\pm$ 0.3  &        15.3 $\pm$ 0.2   &       70.3 $\pm$ 0.1     &     55.9 $\pm$ 0.2  & 48.1 $\pm$ 0.1  \\
ERM-SDViT & T2T-ViT-14 & 21.5M & 67.6 $\pm$ 0.2 &     25.0 $\pm$ 0.2&    55.8 $\pm$ 0.4   &       15.2 $\pm$ 0.3&     70.0 $\pm$ 0.1   &       55.9 $\pm$ 0.1  & 48.2 $\pm$ 0.2 \\
ERM-SDViT + T3A & T2T-ViT-14 & 21.5M & 68.2 $\pm$ 0.1       &   25.8 $\pm$ 0.2    &      56.7 $\pm$ 0.3       &   20.7 $\pm$ 0.2    &      72.4 $\pm$ 0.1       &   57.0 $\pm$ 0.2        & 50.2 $\pm$ 0.1    \\
\hline
\end{tabular}
}
\end{center}
\end{table}
\noindent\textbf{Training overhead on target domains of other DG datasets:} Table~\ref{comp_time_Terra} and \ref{comp_time_DomainNet} reports training overhead, computed as relative \% increase in training time (hrs.) on \texttt{TerraIncognita} and \texttt{DomainNet} datasets. The numbers report the training time increase introduced by our method on top of the baseline. The results show that in both large-scale DG benchmark datasets i.e. \texttt{TerraIncognita} (24K images) and \texttt{DomainNet} (500K images), our model (ERM-SDViT) is not exceeding more than 20\% relative overhead training time. Note that this training time could differ with GPU utilization. Results are reported with DeiT-Small (22M params.) backbone on Nvidia RTX A6000 GPU.

\begin{table}[!t]
\caption{\small Training overhead, computed as relative \% increase in training time (hrs.), introduced by our method on top of the baseline.}
\resizebox{\textwidth}{!}{
\begin{tabular}{lcccc}
\hline \noalign{\smallskip}
\rowcolor{Gray}
Dataset: & & \texttt{TerraIncognita}& & \\
\hline
\rowcolor{Gray}
Model  & Location\_38 & Location\_43 & Location\_46 & Location\_100  \\
\hline
ERM-ViT  &  0.268&	0.268&	0.270	&0.268 	  \\
ERM-SDViT  & 0.276	&0.282&	0.282&	0.302 \\
\hline
Rel.overhead & 2.975&	5.068&	4.447&	12.620  \\
\noalign{\smallskip}
\hline
\end{tabular}
}
\label{comp_time_Terra}
\vspace{-0.5em}
\end{table}
\begin{table}[!t]
\caption{\small Training overhead, computed as relative \% increase in training time (hrs.), introduced by our method on top of the baseline.}
\resizebox{\textwidth}{!}{
\begin{tabular}{lcccccc}
\hline \noalign{\smallskip}
\rowcolor{Gray}
Dataset: &  & & \texttt{DomainNet} &  & &\\
\hline
\rowcolor{Gray}
Model  & Clipart & Infograph & Painting & Quickdraw & Real & Sketch   \\
\hline
ERM-ViT  & 0.418&	0.423&	0.422&	0.430&	0.436 &	0.430	  \\
ERM-SDViT  & 0.482	&0.444&	0.510	&0.469&	0.446&	0.460 \\
\hline
Rel.overhead & 15.376&	5.124&	20.928	&9.079&	2.463&	7.089
  \\
\hline
\end{tabular}
}
\label{comp_time_DomainNet}
\vspace{-1.5em}
\end{table}


%
\end{document}